\documentclass[10pt,twocolumn,letterpaper]{article}

\usepackage[pagenumbers]{cvpr} 
\makeatletter
\@namedef{ver@everyshi.sty}{}
\makeatother
\usepackage{tikz}
\usepackage{graphicx}
\usepackage{amsmath}
\usepackage{amssymb}
\usepackage{booktabs}
\usepackage{tikz}
\usepackage{comment}
\usepackage{color}
\usepackage{multirow}
\usepackage{kotex}
\usepackage{verbatim}
\usepackage{makecell}
\usepackage{caption}
\usepackage{subcaption}
\usepackage{cite} 

\usepackage{amsmath,amsfonts,amssymb}
\usepackage{bm}
\usepackage{nicefrac}
\usepackage{microtype}
\usepackage{dsfont}
\usepackage{changepage}
\usepackage{extramarks}
\usepackage{fancyhdr}
\usepackage{setspace}
\usepackage{soul}
\usepackage{xspace}
\usepackage{booktabs}
\usepackage{adjustbox}

\usepackage[pagebackref,breaklinks,colorlinks]{hyperref}

\newcommand{\Secref}[1]{\textbf{Section~\ref{sec:#1}}}
\newcommand{\figref}[1]{\textbf{Fig.~\ref{fig:#1}}}
\newcommand{\tabref}[1]{\textbf{Table~\ref{tab:#1}}}

\newcommand{\secref}[1]{Sec.~\ref{sec:#1}}

\newcommand{\ours}{GGDR~\xspace{}}

\newcolumntype{x}[1]{>{\centering\arraybackslash\hspace{0pt}}p{#1}}

\usepackage[capitalize]{cleveref}
\crefname{section}{Sec.}{Secs.}
\Crefname{section}{Section}{Sections}
\Crefname{table}{Table}{Tables}
\crefname{table}{Tab.}{Tabs.}

\def\cvprPaperID{*****} 
\def\confName{CVPR}
\def\confYear{2022}

\begin{document}

\title{Generator Knows What Discriminator Should Learn in Unconditional GANs}

\author{Gayoung Lee$^1$ \quad Hyunsu Kim$^1$ \quad Junho Kim$^1$ \\
Seonghyeon Kim$^2$  \quad Jung-Woo Ha$^1$ \quad Yunjey Choi$^1$ \\ \\
$^{1}$NAVER AI Lab, $^2$NAVER CLOVA\\
\tt\small \{gayoung.lee, hyunsu1125.kim, jhkim.ai,\\\tt\small kim.seonghyeon, jungwoo.ha, yunjey.choi\}@navercorp.com}

\maketitle

\begin{abstract}
Recent methods for conditional image generation benefit from dense supervision such as segmentation label maps to achieve high-fidelity. However, it is rarely explored to employ dense supervision for unconditional image generation. Here we explore the efficacy of dense supervision in unconditional generation and find generator feature maps can be an alternative of cost-expensive semantic label maps. From our empirical evidences, we propose a new \textit{generator-guided discriminator regularization} (\textit{GGDR}) in which the generator feature maps supervise the discriminator to have rich semantic representations in unconditional generation. In specific, we employ an U-Net architecture for discriminator, which is trained to predict the generator feature maps given fake images as inputs. Extensive experiments on mulitple datasets show that our GGDR consistently improves the performance of baseline methods in terms of quantitative and qualitative aspects. Code is available at \url{https://github.com/naver-ai/GGDR}.
\end{abstract}
\section{Introduction}
\label{sec:intro}

Generative adversarial networks(GANs) have achieved promising results in various computer vision tasks including image~\cite{karras2019stylegan,karras2020stylegan2,karras2021alias} or video generation~\cite{skorokhodov2021stylegan,yu2022generating,tian2021good,tulyakov2018mocogan}, translation~\cite{isola2017image,zhu2017cyclegan,choi2018stargan,kim2019tag2pix,Kim2020U-GAT-IT}, manipulation~\cite{bau2018gan,kim2021exploiting,jahanian2019steerability,shen2020interfacegan,harkonen2020ganspace,kim2022contrastive}, and cross-domain translation~\cite{kim2017learning,huang2018auggan} for the past several years.
In GANs, building an effective discriminator is one of the key components for generation quality since the generator is trained by the feedback from the discriminator. Existing studies proposed various methods to make the discriminator learn better representations by data augmentation~\cite{karras2020ada,zhao2020differentiable,zhang2019consistency,zhao2020improved}, gradient penalty~\cite{mescheder2018r1reg,miyato2018spectral,roth2017stabilizing}, and carefully designed architectures~\cite{schonfeld2020u,karras2017progressive}.

One simple yet effective way to improve the discriminator is to provide available additional annotations such as class labels~\cite{odena2017conditional,brock2018large}, pose descriptors~\cite{siarohin2018deformable}, normal maps~\cite{wang2016normalmap}, and semantic label maps~\cite{isola2017image,park2019spade,sushko2020oasis,liu2019ccfpse}. Among these annotations, semantic label maps contain rich and dense descriptions about images, and have been frequently used in 
conditional scene generation. To provide dense semantic information to the discriminator, Pix2pix~\cite{isola2017image} and SPADE~\cite{park2019spade} concatenate the label maps with input images, and CC-FPSE~\cite{liu2019ccfpse} uses projection instead of the concatenation to inject the embedding of label maps. OASIS~\cite{sushko2020oasis} further enhances the discriminator by providing strong supervision using auxiliary semantic segmentation task and achieves better performance.

\begin{figure}[t]
  \centering
\includegraphics[width=1.0\linewidth]{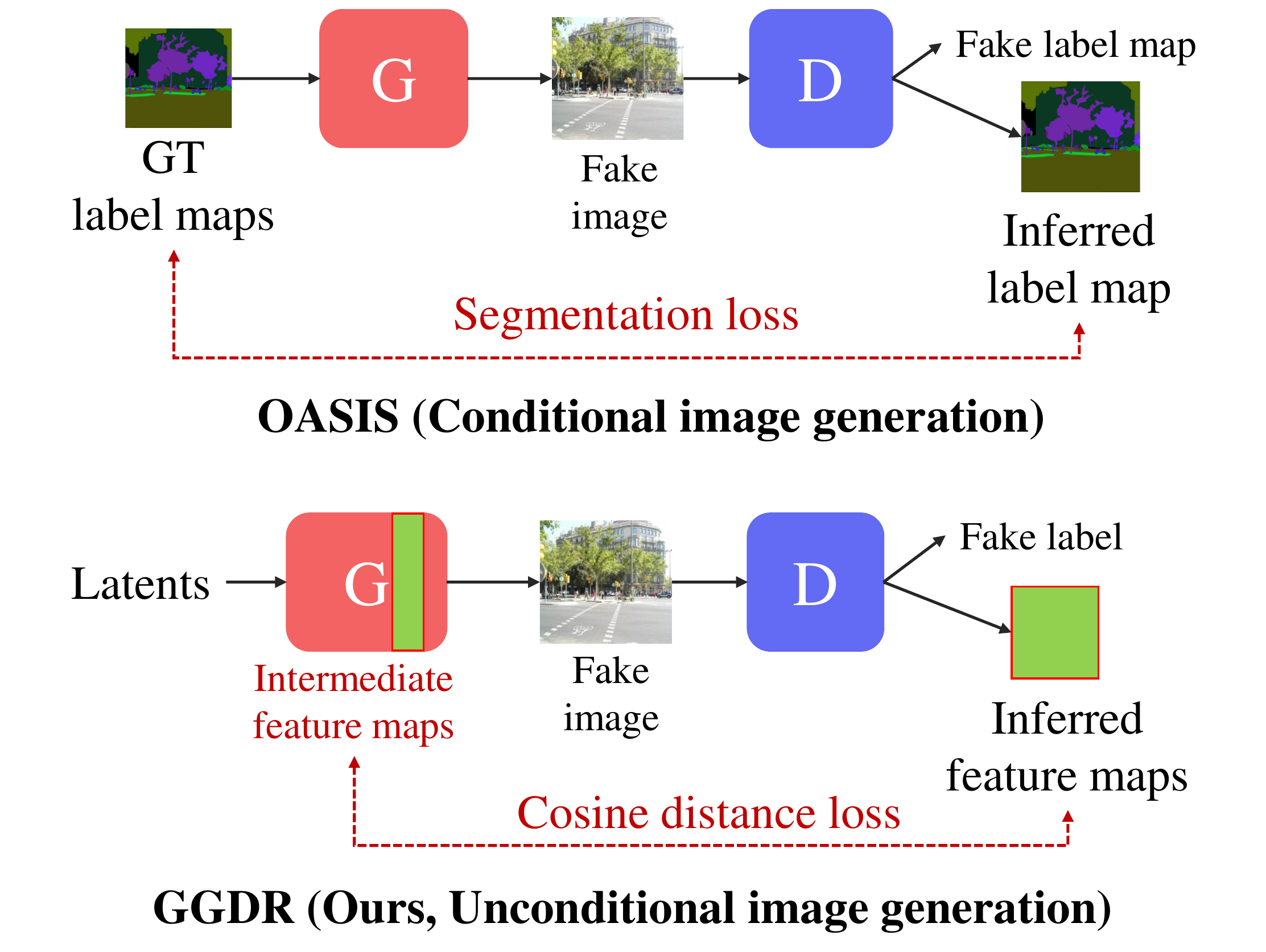}
  \caption{Comparison of how to provide semantic information between OASIS and our method. OASIS enhance the discriminator with the ground truth label maps in conditional image generation setting. GGDR, on the other hand, aims at unconditional image synthesis, and uses the generator feature maps instead of human-annotating label maps.}
  \label{fig:cmp_oasis}
\end{figure}

Despite the success of dense semantic supervision in conditional generation, it has been rarely explored in an unconditional setting. Dense semantic supervision can be useful here as well, as GAN models often struggle when the data has varied and complex layout images. However, in unconditional generation, most large datasets do not have pairs of images and semantic label maps, since collecting them has a significant human annotation cost. Therefore, unlike the conditional setting, which requires a dense label map for the generator input, unconditional image generation assumes no dense map, and most studies use discriminators that learn only from images.

In this paper, we show that guiding a discriminator using dense and rich semantic information is also useful in unconditional image generation, and propose the method that avoids data annotation costs while utilizing semantic supervision. We propose \textit{generator-guided discriminator regularization} (\textit{GGDR}) in which the generator feature maps supervise the discriminator to have rich semantic representations. Specifically, we redesign the discriminator architecture in U-Net style, and train the discriminator to estimate the generator feature map when input is a generated image. As shown in \figref{cmp_oasis}, GGDR differs from the previous work in that the discriminator is supervised by the generator feature maps instead of human-annotated semantic label maps.   

To justify our proposed method, we first compare the generation performance of StyleGAN2~\cite{karras2020stylegan2} with and without providing ground-truth segmentation maps to the discriminator, and show that utilizing semantic label maps indeed improves the generation performance in an unconditional setting (\Secref{semantic_label_map}). We then visualize the generator feature maps and show that they contain semantic information rich enough to guide the discriminator, replacing the ground-truth label maps (\Secref{analysis_feature_maps}). Utilizing the generator feature maps, GGDR improves the discriminator representation, which is the key component to enhance the generation performance (\Secref{method}). We provide thorough comparisons to demonstrate that GGDR consistently improves the baseline models on a variety of data. Our method can be easily attached to any setting without burdensome cost; only 3.7\% of the network parameter increased. Our contributions can be summarized as follows:

\begin{enumerate}
	\item We investigate the effectiveness of dense semantic supervision on unconditional image generation.
	\item We show that generator feature maps can be used as an effective alternative of human-annotated semantic label maps.
	\item We propose generator-guided discriminator regularization (GGDR), which encourages the discriminator to have rich semantic representation by utilizing generator feature maps.
	\item We demonstrate that GGDR consistently improves the state-of-the-art methods on multiple datasets, especially in terms of generation diversity. 
	
\end{enumerate}
\section{Dense semantic supervision in unconditional GANs}
\label{sec:analysis}

\begin{figure}[t]
\centering
    \includegraphics[width=0.8\linewidth]{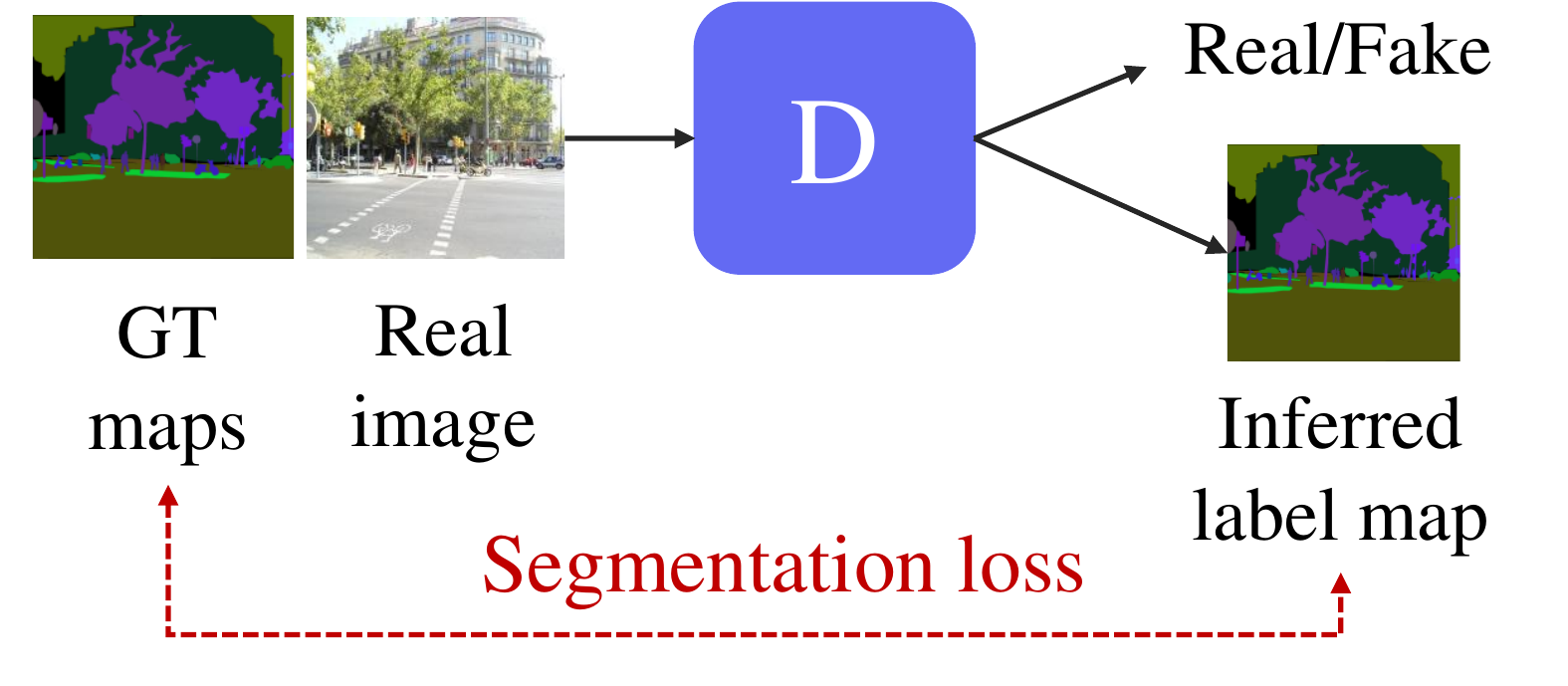}
    \caption{Discriminator architecture with the auxiliary segmentation loss for the preliminary experiments.}
  \label{fig:ade20k_darch}
\end{figure}
\begin{figure}
\centering
    \centering
    \includegraphics[width=0.8\linewidth]{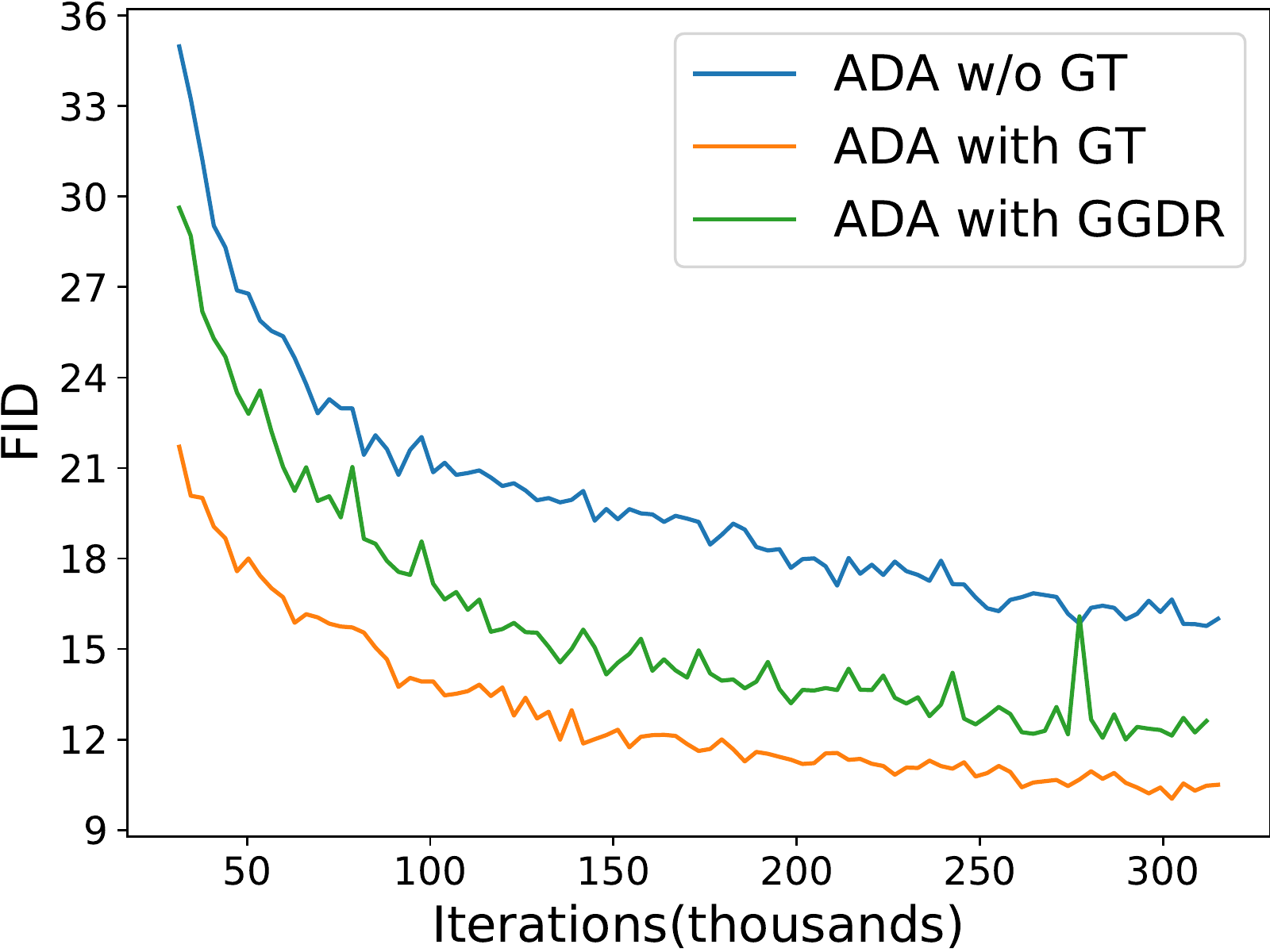}
    \caption{FID scores on ADE20K with and without dense semantic supervision in the preliminary experiments.}
  \label{fig:ade20k_gfeat}
\end{figure}

We first conduct a preliminary experiment using ground-truth segmentation maps to show the efficacy of providing dense semantic supervision for the discriminator (\Secref{semantic_label_map}). Then, we study whether the generator feature maps can be used as a guide instead of using human annotating ground-truth label maps to avoid expensive manual annotations. We visualize the internal feature maps of the generator and show that they have semantic information rich enough to be used as pseudo-semantic labels (\Secref{analysis_feature_maps}). 

\subsection{Utilization of semantic label maps for discriminator} 
\label{sec:semantic_label_map}

Although it is natural to utilize semantic label maps in conditional image generation~\cite{park2019spade,liu2019ccfpse,sushko2020oasis}, it has been still underexplored whether label maps are beneficial for unconditional image generation~\cite{karras2019stylegan,karras2020stylegan2,karras2020ada}. We conduct a preliminary experiment to validate the effects of the semantic label maps. We use ADE20K scene parsing benchmark dataset~\cite{zhou2017scene} consisting of 20,210 paired images and semantic label map annotations with 150 class labels, which is frequently used to evaluate conditional generation models. We choose StyleGAN2~\cite{karras2020stylegan2} as our baseline, which is a standard model for unconditional image generation and apply adaptive discriminator augmentation~\cite{karras2020ada}. To provide semantic supervision for the network, we redesign the discriminator to perform additional segmentation task similar to OASIS~\cite{sushko2020oasis}. The modified task for the discriminator is described in \figref{ade20k_darch}. The detailed architecture is similar to \figref{model}, except that it upsamples the decoder output until the image size is reached. The decoder in a U-Net style is attached to the discriminator and the segmentation loss is applied to the last layer of the decoder to provide dense supervision. The segmentation loss is the usual cross-entropy loss. Since ground-truth label maps are not available for generated images, we activate the segmentation loss for real images only.

 As shown in \figref{ade20k_gfeat}, the model with the discriminator leveraging the semantic supervision(ADA with GT) outperforms the baseline(ADA w/o GT). As argued in OASIS, the stronger semantic supervision seems to help discriminator learn more semantically and spatially-aware representations and give the generator more meaningful feedback. Our experiment supports that providing additional semantic guide for the discriminator can improve the model performance in unconditional image synthesis. However, dense label maps are rare in datasets for unconditional image synthesis, and it is time-consuming to collect them manually. In the next section, we analyze the feature maps from the generator as an effective alternative for the ground-truth label maps.

\subsection{Analysis of generator feature maps}
\label{sec:analysis_feature_maps}

\begin{figure}[t]
\centering

\begin{subfigure}[t]{0.9\linewidth}
\centering
    \includegraphics[width=1.0\textwidth]{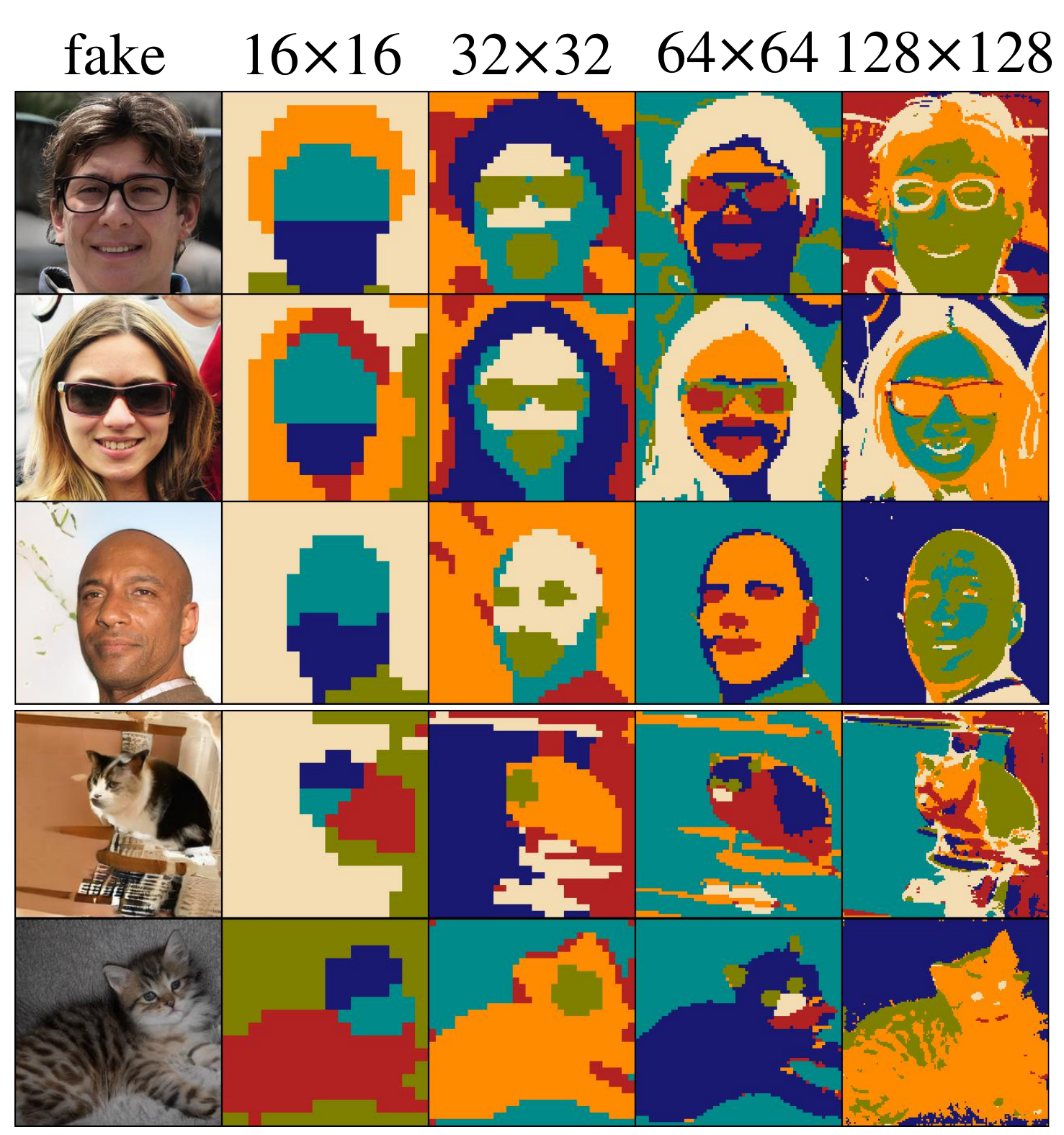}
\caption{Generator feature map visualization}
\hspace{0.5cm}
  \end{subfigure}
\begin{subfigure}[t]{1.0\linewidth}
\centering
    \includegraphics[width=1.0\linewidth]{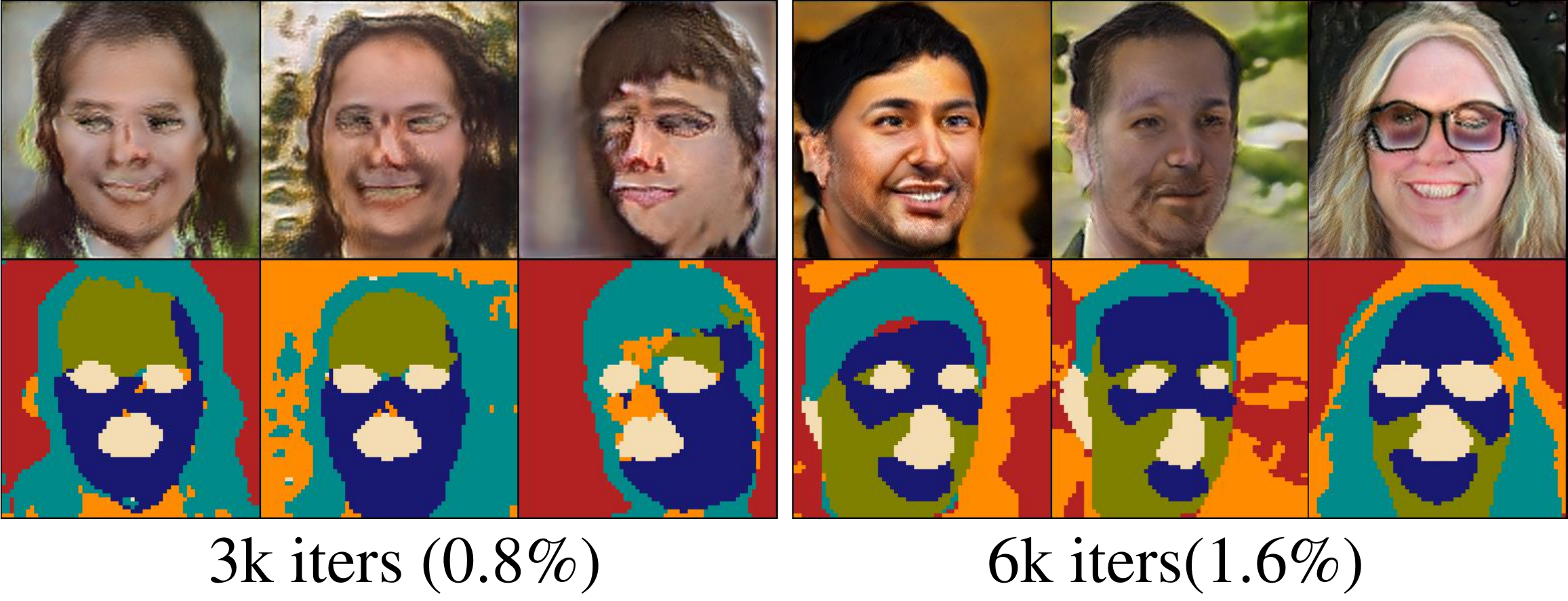}
    \caption{Generator feature map in early training phase}
  \end{subfigure}
  \caption{Visualization of the generator feature maps using $k$-means($k=6$) clustering. (a) StyleGAN2 generator feature map. The visualized feature maps reveal semantically consistent and meaningful regions such as ears in cats. (b) Generated images and their $32 \times 32$ feature maps in the early training phase.}
   \label{fig:g_feats}
\end{figure}

Recent studies have reported that the feature maps of the trained generator of GANs contain rich and dense semantic information~\cite{collins2020editing,xu2021linear,endo2021fewshotsmis}. Collins et al.\cite{collins2020editing} showed that applying $k$-means clustering to the feature maps of the generator reveals semantics and parts of objects, and used the clusters to edit images. We notice that these feature maps are rich semantic descriptors of the generated images and can be the substitute for the ground truth label maps. To visualize what information is captured in each feature map, we run $k$-means algorithm on each layer using the batch of generated images. We set $k=6$ in this experiment. As shown in \figref{g_feats} (a), the pixels are clustered by the semantic information instead of the low-level features except the last feature map. For example, the hairs of the people have different colors, but are clustered in to the same cluster. The early feature maps show coarse object location, and those from the latter layers contain detailed object parts. The visualized feature maps look like pseudo-semantic label maps and might be regarded as rich descriptions including spatial and semantic information about the images. Therefore, we choose the feature maps of the generator as the substitute for the semantic label maps to guide the discriminator using semantic supervision. The generator feature maps are useful in our case. First, we do not need perfect semantic segmentation maps because our goal is image generation not semantic segmentation. Second, the feature maps are intermediate by-products essential for the generation, so acquiring them is free and does not require additional human annotations. 

Dissimilar to previous works~\cite{collins2020editing,xu2021linear,endo2021fewshotsmis} that utilize the generator feature map for separate tasks, our method utilizes them during the training to enhance the generation performance itself. Therefore, it is essential to check whether the feature maps from the generator in the middle of training are still semantically meaningful for the guidance. In \figref{g_feats} (b), we visualize the feature maps of the generator during training to check how early the feature maps become semantically meaningful. Surprisingly, thanks to the powerful modern GANs, we can observe that even in the early stage, the feature maps and the corresponding generated images capture coarse shapes and location of objects. Therefore, we utilize the feature maps from the beginning of training, but for more complex data where the generator needs more iterations to produce meaningful semantics, one may choose when to attach our objective function.
\section{Generator-guided discriminator regularization}
\label{sec:method}

\begin{figure}[t]
\centering
\begin{subfigure}[t]{0.9\linewidth}
\centering
\includegraphics[width=\textwidth]{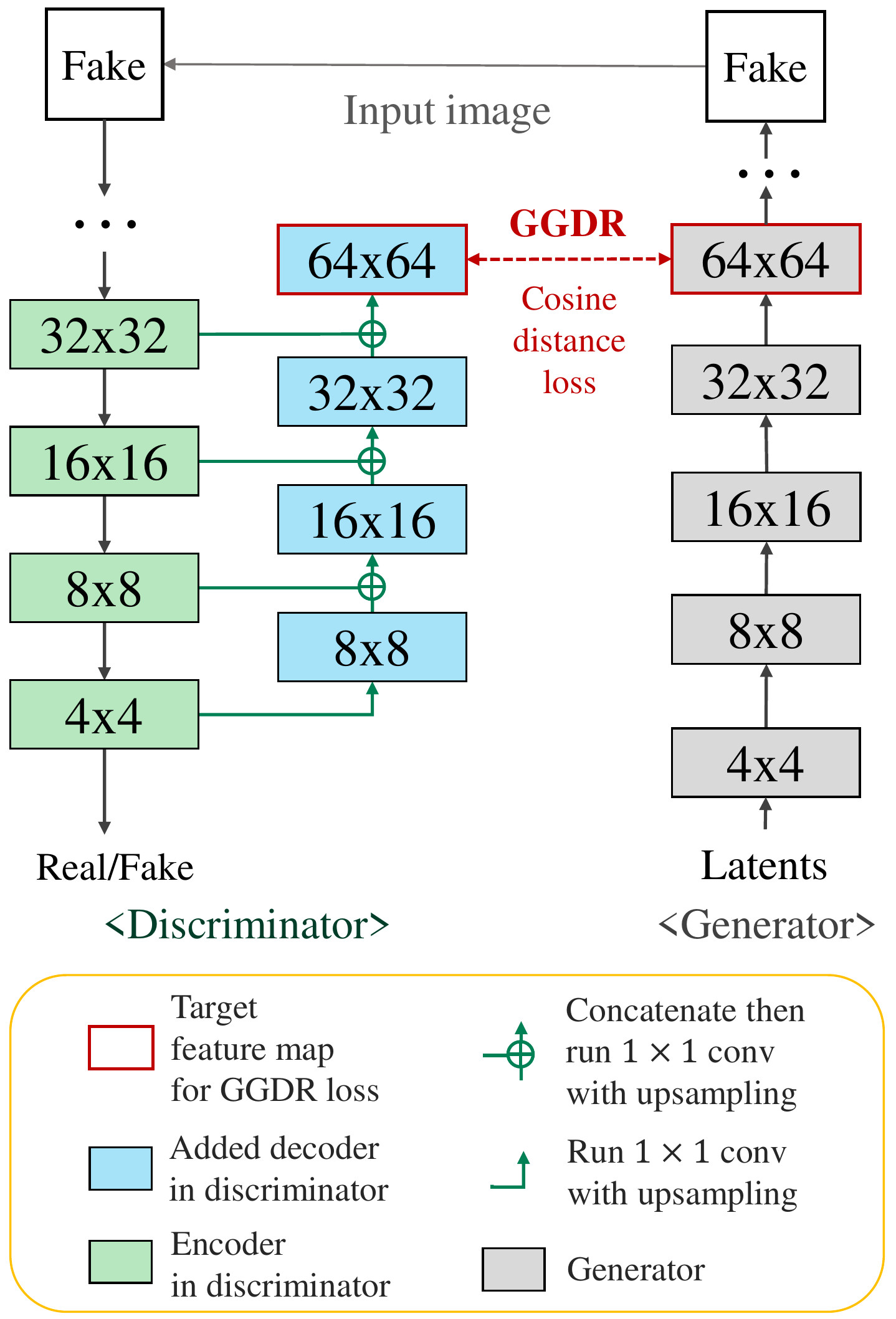}
\end{subfigure}
\caption{Visualization of our framework. Our method can be applied to a GAN model by adding a decoder and the cosine distance loss with the reference generator feature maps to a discriminator. The generator feature maps guide the discriminator to learn more semantically-aware representations.}
\label{fig:model}
\end{figure}

Based on our observations, we propose \textit{generator-guided discriminator regularization} (GGDR) in which the generator feature maps supervise the discriminator to have rich semantic representations. The overall framework is shown in  \figref{model}. 

The design of our discriminator $D$ is inspired by that of OASIS~\cite{sushko2020oasis} where the U-Net encoder-decoder structure is adopted and the last layer predicts semantic label maps. However, unlike OASIS, we leverage feature maps of the generator instead of ground-truth label maps. Thus, there are several differences in the design. First, since the feature maps are not discrete labels anymore, we cannot simply add real/fake class to the decoder output as done in OASIS. Therefore, we separate the decoder and adversarial loss. Next, we use more compact and lighter modules to reduce additional calculation costs. For each layer, we concatenate the output from the decoder and the encoder layer, and run one linear $1\times1$ convolutional layer with upsampling. We stack the decoder modules until the decoder output has the same resolution with the targeted generator feature map. Although the decoder is compact, it is sufficient to predict the generator feature map as the shared encoder can extract semantic information. 

Meanwhile, the encoder part is still shared, and thus it is trained via both semantic and adversarial loss. For the adversarial loss, we adopt the non-saturating adversarial loss~\cite{goodfellow2014gan}: 
\def\L{\mathcal{L}} %
\newcommand{\expect}[1]{\mathbb{E}_{#1}}
\def\x{\bm{x}} %
\def\z{\bm{z}} %
\begin{multline}
\min_{G}\max_{D}\L_{adv}(G, D) = \mathbb{E}_{\x\sim p_{data}(\x)}[\log D(\x)] + \\
\mathbb{E}_{\z \sim p(\z)}[\log(1-D(G(\z)))]
\end{multline}

Next, we compute the cosine distance loss between the output of the decoder and the target feature map. We use cosine distance loss since it gives a loss within a specified range even between denormalized feature vectors, so it is convenient to scale according to the adversarial loss. Here, we denote $l\in \{1,2,...,L\}$ as a layer index, where $L$ is the number of decoding layers. Our discriminator $D$ contains an U-Net-style decoder $F$ ($F \subset D$) and the output of each layer denoted as $F^{l}$ has the same resolution with the corresponding generator feature map $G(\z)^l$. Let us denote the target layer index for guidance as $t$. Our generator-guided discriminator regularization (GGDR) is defined as:

\begin{multline}
\max_{D}\L_\textit{ggdr}(G, D) =\\ -\expect{\z \sim p(\z)} \Bigg[ 1 - \frac{F^{t}(G(z))\cdot G(z)^{t}}{\left\| F^{t}(G(z)) \right\|_2\cdot \left\| G(z)^{t} \right\|_2} \Bigg]
\label{eq:nce_int}
\end{multline}

Our full objective functions can be summarized as follows:

\begin{equation}
\L_{total}  = \L_{adv} + \lambda_{reg}\L_{ggdr}
\end{equation}

\noindent where $\lambda_{reg}$ is a hyperparameter for relative strength compared to the adversarial term. Note that $k$-means clustering used in \Secref{analysis_feature_maps} is only for visualization purpose and we directly compare raw feature maps without any clustering. We expect the $\L_{ggdr}$ term to enhance the semantic representation of $D$. While the generator feature maps participate the regularization loss, we do not update the generator with $\L_{ggdr}$ to prevent a feature collapse which is a trivial solution making the cosine distance to zeros. 

Our framework is simple and easy to apply existing GAN models, and does not require any additional annotation. Acquiring the intermediate feature maps from a generator is free because the generator already produces them in order to generate fake images. Despite its simplicity, in the next section, we show the effectiveness of our method in unconditional image generation for various datasets.

\section{Experiments}
\label{sec:exp}

We validate the efficacy of our \ours on various datasets including CIFAR-10~\cite{krizhevsky2009cifar10}, FFHQ~\cite{karras2019stylegan}, LSUN cat, horse, church~\cite{yu2015lsun}, AFHQ~\cite{choi2020stargan} and Landscapes~\cite{landscape2019kaggle}. CIFAR-10 consists of 50,000 tiny color images in 10 classes. FFHQ contains 70,000 face images, and AFHQ includes approximately 5,000 images per cat, dog and wild animal faces. LSUN cat, horse and church consist of scenes with cat, horse and church respectively, and we have used 200,000 images per each dataset. Landscapes contains photographs 4,320 landscape images collected from Flickr~\cite{landscape2019flickr}. Following StyleGAN2 and ADA~\cite{karras2020ada}, we have applied horizontal flips for FFHQ and small datasets. All images are resized to $256 \times 256$ except AFHQ ($512 \times 512$) and CIFAR-10 ($32 \times 32$). For GGDR loss, we select the $64 \times 64$ feature map of the generator as the guidance map, except CIFAR-10 where we select the $8 \times 8$ feature map. We set $\lambda_{reg} = 10$ for the weight of the proposed regularization in all experiments and let other hyperparameters unmodified. We apply ${R}_{1}$ regularization~\cite{mescheder2018r1reg} for StyleGAN2 and ADA models. In the case of the ADA, we apply the augmentation to the generator feature maps to make them consistent with corresponding fake images. We use only geometric operations and skip color transformation for the feature map augmentation.

For evaluation metrics, we have used Fr\'{e}chet Inception Distance(FID)~\cite{heusel2017fid} and Precision \& Recall~\cite{kynkaanniemi2019improved}. FID measures the distance between the real images and the generate samples in feature space, and Precision \& Recall scores indicate sample quality and variety. We compare 50,000 generated images and all training images following previous works~\cite{karras2020ada}. For CIFAR-10, we also use Inception Score (IS)~\cite{salimans2016improved} following the previous works~\cite{karras2020ada, song2020score}.

\subsection{Comparison with baselines} 
\newcommand{\std}[1]{\makebox[7.3mm][r]{\footnotesize\raisebox{0.3mm}{\scalebox{0.9}{$\pm$}}\hspace{0.2mm}#1}}
\newcommand{\maketiny}[1]{\makebox[6.3mm][r]{\scriptsize #1}}
{
\begin{table}[t]
\centering
\caption{FID scores of ours and comparison methods on FFHQ. We run three training for each data and show their means and standard deviations. The numbers are largely brought from ADA~\cite{karras2020ada} and we follow their evaluation protocol.}
\label{tab:sota_ffhq_table}
\begin{tabular}[t]{rccc}
\toprule
FFHQ & 2k & 10k & 140k \\
 \midrule
PA-GAN & 56.49\std{7.28} & 27.71\std{2.77} & 3.78\std{0.06} \\

WGAN-GP & 79.19\std{6.30} & 35.68\std{1.27} & 6.43\std{0.37}  \\

zCR  & 71.61\std{9.64} & 23.02\std{2.09} & 3.45\std{0.19} \\

AR  & 66.64\std{3.64} & 25.37\std{1.45} & 4.16\std{0.05} \\

 \midrule
StyleGAN2 & 78.80\std{2.31} & 30.73\std{0.48} & 3.66\std{0.10}  \\
+GGDR & 70.59\std{5.16} & 24.44\std{0.63} & \textbf{3.14\std{0.03}} \\
 \midrule
ADA & \textbf{16.49\std{0.65}}& 8.29\std{0.31} & 3.88\std{0.13} \\
+GGDR & 18.28\std{0.77} & \textbf{6.11\std{0.15}} & 3.57\std{0.10} \\
\bottomrule
\end{tabular}
\end{table} 

\begin{table}[t]
\centering
\caption{FID scores of ours and comparison methods on CIFAR-10. We run three training for mean and standard deviations. We brought the numbers of diffusion models from ~\cite{song2020score}.}
\label{tab:sota_cifar10_table}
\begin{tabular}[t]{ r c c}
\toprule
CIFAR-10 & FID & IS \\
 \midrule
ProGAN & 15.52 & 8.56\std{0.06} \\ 
AutoGAN & 12.42 & 8.55\std{0.10} \\ 
StyleGAN2 & 8.32\std{0.09} & 9.21\std{0.09} \\
ADA & 2.92\std{0.05} & 9.83\std{0.04} \\
FSMR & 2.90 & 9.68 \\
\midrule
DDPM & 3.17\std{0.05} & 9.46\std{0.11} \\
NCSN++ & 2.2 & 9.89 \\
\midrule
ADA+\ours & \textbf{2.15\std{0.02}} & \textbf{10.02\std{0.06}} \\
\bottomrule
\end{tabular}
\end{table} 
}

We apply GGDR to StyleGAN2~\cite{karras2020stylegan2} which is one of the standard models for unconditional image generation. Instead of the original StyleGAN2 setting, we use the baseline setting used in ADA~\cite{karras2020ada} which has less parameters and shorter training iterations but shows comparable performance. For small datasets, we apply the adaptive discriminator augmentations (ADA)~\cite{karras2020ada} that prevents overfitting of the discriminator and shows the superior performance in small datasets. 

\tabref{sota_ffhq_table} and \tabref{sota_cifar10_table} shows that the proposed GGDR improves the performance of the baselines in terms of FID scores which indicate the overall quality of the synthesized images. Following ADA~\cite{karras2020ada}, we run the experiment multiple times on FFHQ with varying numbers (2k, 10k and 140k) of training images. We largely borrowed the reported scores from ADA~\cite{karras2020ada} and NCSN++~\cite{song2020score}. In \tabref{sota_ffhq_table}, we compare our method in a varied number of training images with various regularizing methods WGAN-GP~\cite{gulrajani2017wgangp}, PA-GAN~\cite{zhang2018pa}, zCR~\cite{zhao2020improved}, AR~\cite{chen2019rotation} and ADA~\cite{karras2020ada}. In the table, StyleGAN2 with our GGDR achieves the best score on FFHQ with full training images. With a sufficient number of training images, regularization methods based on data augmentation show limited improvement or degradation, whereas in our method the discriminator learns from a more accurate semantic map provided by a better quality generator. In most cases, GGDR improves the baseline performance significantly except the FFHQ 2k setting. We conjecture the dataset is too small to learn semantically meaningful feature maps in the generator. Since the quality of generator feature maps directly affects the discriminator in our method, our method is more effective when there is sufficient number of training data. However, as shown in \tabref{sota_comparison_small1}, our method shows effectiveness on the datasets with approximately 5,000 images which are not too many to collect. \tabref{sota_cifar10_table} shows that our GGDR significantly improves ADA performance in terms of both FID and IS scores, and makes it superior to various models including ProGAN~\cite{karras2017progressive}, AutoGAN~\cite{gong2019autogan}, FSMR~\cite{kim2021feature} and DDPM~\cite{ho2020denoising}, and comparable to the NCSN++~\cite{song2020score}.

{
\begin{table*}[t]
\begin{center}
\caption{Comparision on FFHQ, LSUN Cat, LSUN Horse and LSUN Church. Our method improves StyleGAN2~\cite{karras2020stylegan2} in large datasets in terms of FID and recall. P and R denote precision and recall. Lower FID and higher precision and recall mean better performance. The bold numbers indicate the best FID, P, R for each dataset.}
\label{tab:sota_comparison_large}
\small{
\begin{tabular}{r c c c c c c c c c c c c}
\toprule
\multirow{2}{*}{Method} & \multicolumn{3}{c}{FFHQ} & \multicolumn{3}{c}{LSUN Cat} & \multicolumn{3}{c}{LSUN Horse} & \multicolumn{3}{c}{LSUN Church} \\
\cmidrule(lr){2-4}
\cmidrule(lr){5-7}
\cmidrule(lr){8-10}
\cmidrule(lr){11-13}
&FID$\downarrow$ &P$\uparrow$ &R$\uparrow$ &FID &P &R &FID &P &R &FID &P &R\\
\midrule
UT~\cite{bond2021unleashing} &  6.11 & \textbf{0.73} & 0.48 &  - & - & - &  - & - & - &  4.07 &  \textbf{0.71} & 0.45 \\
Polarity~\cite{humayun2022polarity} & - & - & - &   6.39 & \textbf{0.64} & 0.32 &  - & - & - &  3.92 &  0.61 & 0.39 \\
\midrule
StyleGAN2 &3.71 &0.69 &0.44 &7.98 & 0.60 &0.27 &3.62 & 0.63 &0.36& 3.97 & 0.59 & 0.39 \\
+\ours & \textbf{3.14} & 0.69 &\textbf{0.50} &\textbf{5.28} &0.58 &\textbf{0.38} &\textbf{2.50} & \textbf{0.64} & \textbf{0.43} & \textbf{3.15} & 0.61 & \textbf{0.46}\\
\bottomrule
\end{tabular}
}
\end{center}
\end{table*}
}
{
\begin{table*}[t]
\begin{center}
\caption{Comparision on AFHQ Cat, Dog, Wild and Landscape. Our method improves ADA~\cite{karras2020ada} in small datasets in terms of FID and recall. P and R denote precision and recall. The bold numbers indicate the best FID, P, and R of the models.}
\label{tab:sota_comparison_small1}
\small{
\begin{tabular}{r c c c c c c c c c c c c}
\toprule
\multirow{2}{*}{Method} & \multicolumn{3}{c}{AFHQ Cat} & \multicolumn{3}{c}{AFHQ Dog} & \multicolumn{3}{c}{AFHQ Wild} & \multicolumn{3}{c}{Landscape} \\
\cmidrule(lr){2-4}
\cmidrule(lr){5-7}
\cmidrule(lr){8-10}
\cmidrule(lr){11-13}
&FID$\downarrow$ &P$\uparrow$ &R$\uparrow$ &FID &P &R &FID &P &R &FID &P &R \\
\midrule
FastGAN~\cite{liu2020fastgan} & 4.69 & \textbf{0.78} & 0.31 & 13.09 & 0.75 & 0.38 & 3.14 & 0.76 & 0.20 &  16.44 & \textbf{0.77} & 0.16 \\
ContraD~\cite{jeong2021contrad} & 3.82 & - & - & 7.16 & - & - & 2.54 & - & - & - & - & - \\
\midrule
ADA &3.55 & 0.77 &0.41 &7.40 & 0.76 &0.48 &3.05 & 0.76 & 0.13 &13.87 & 0.72 & 0.20 \\
+\ours & \textbf{2.76} & 0.74 & \textbf{0.52} & \textbf{4.59} & \textbf{0.79} & \textbf{0.53} & \textbf{2.06} & \textbf{0.80} &  \textbf{0.27} & \textbf{10.38} & 0.69 & \textbf{0.29} \\
\bottomrule 
\end{tabular}
}
\end{center}
\end{table*}
}

In \textbf{Tables~\ref{tab:sota_comparison_large}} and\textbf{~\ref{tab:sota_comparison_small1}}, we conduct extensive experiments to validate the performance improvement using GGDR on various datasets. For FFHQ and LSUN datasets, we report the scores of UT~\cite{bond2021unleashing}, and Polarity~\cite{humayun2022polarity} which are the state-of-the-arts models that show the improvement on these datasets. For AFHQ and Landscape datasets, we report the score of ContraD~\cite{jeong2021contrad} and FastGAN~\cite{liu2020fastgan} that show significant improvements on image synthesis with small size datasets. We brought the numbers from their papers except FastGAN whose scores are brought from ProjGAN~\cite{sauer2021projected}. GGDR consistently improves the baseline in terms of FID scores with large gap. In terms of precision and recall metrics, GGDR improves the recall with significant margins compared to the baselines, which indirectly indicates where the advantages of our method come from. Better recall scores mean that our model generates more diverse images and less prone to the mode collapse. As it is known that incorporating image-level labels to the discriminator enhances coverage of classes in the data, utilizing pseudo dense semantic information could facilitate semantic diversities in the generated images.

{
\setlength{\tabcolsep}{0pt}
\renewcommand{\arraystretch}{0.5}
\begin{figure}[t]
  \centering
\begin{subfigure}[t]{0.9\linewidth}
\centering
\begin{tabular}[b]{ccccccc}
\multicolumn{3}{c}{ADA} &  \hphantom{ab} &  \multicolumn{3}{c}{ADA+GGDR} \\ \\
\includegraphics[width=0.15\linewidth]{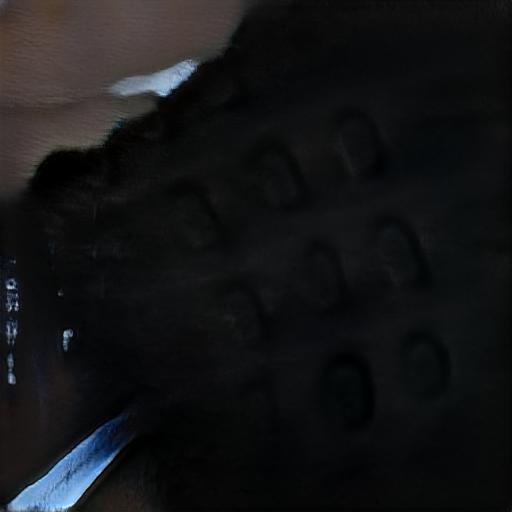} &
\includegraphics[width=0.15\linewidth]{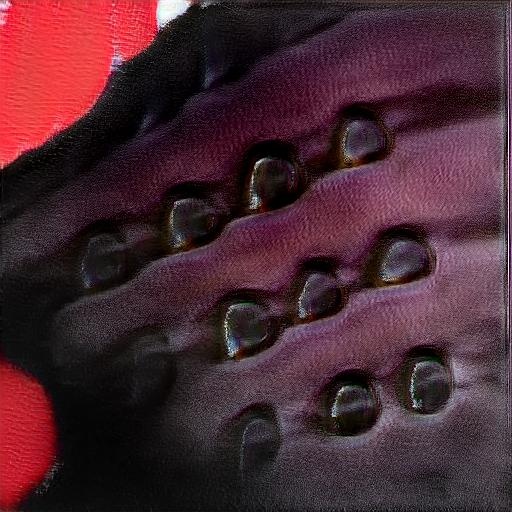} &
\includegraphics[width=0.15\linewidth]{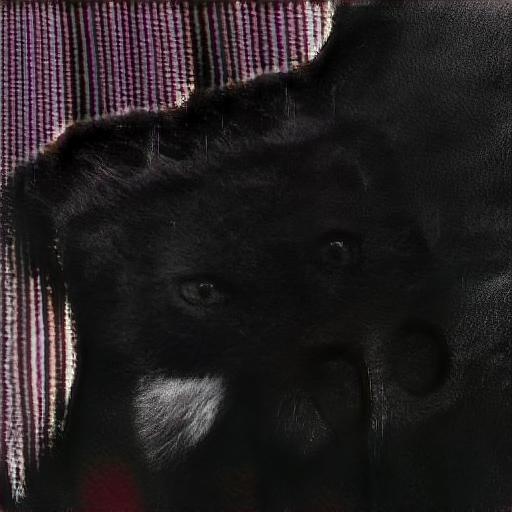} &
 \hphantom{a} &
\includegraphics[width=0.15\linewidth]{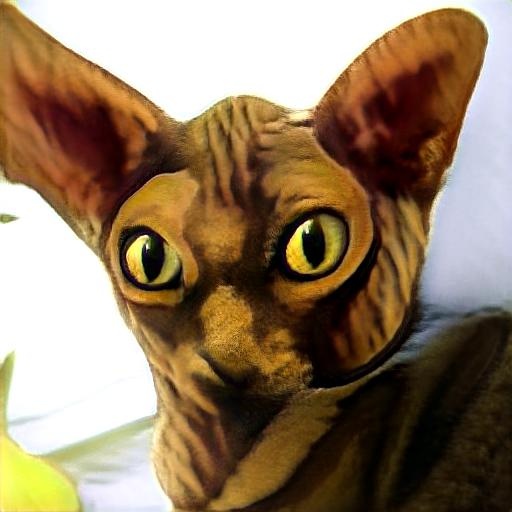} &
\includegraphics[width=0.15\linewidth]{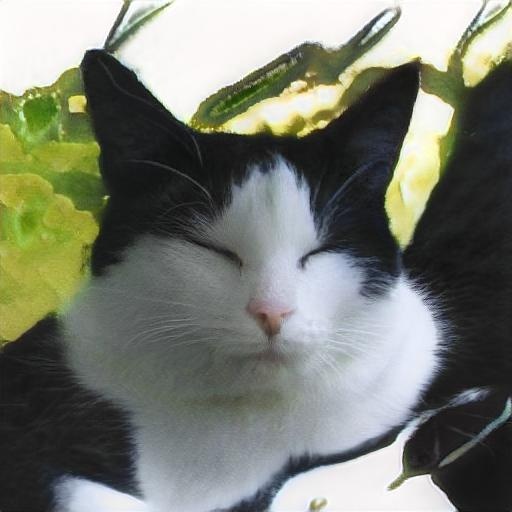} &
\includegraphics[width=0.15\linewidth]{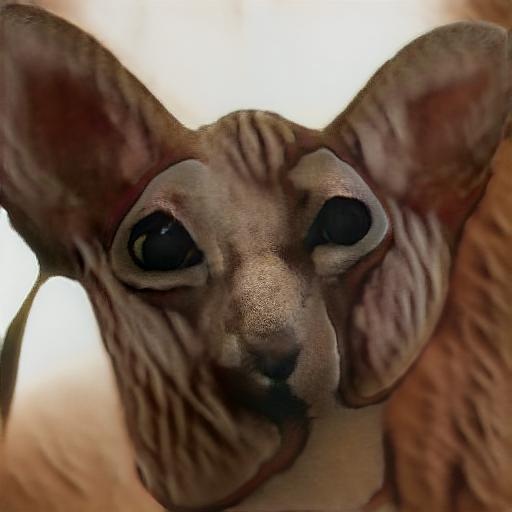} \\

\includegraphics[width=0.15\linewidth]{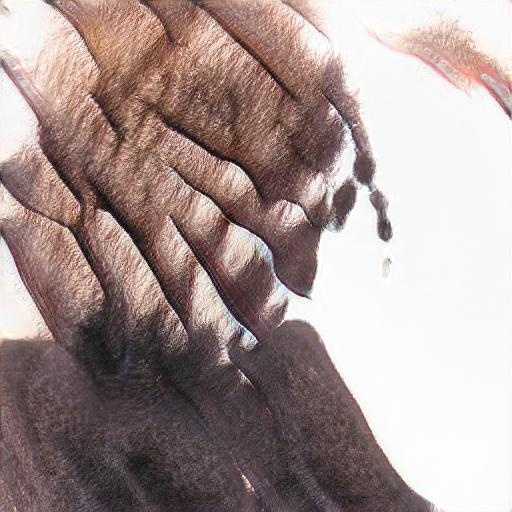} &
\includegraphics[width=0.15\linewidth]{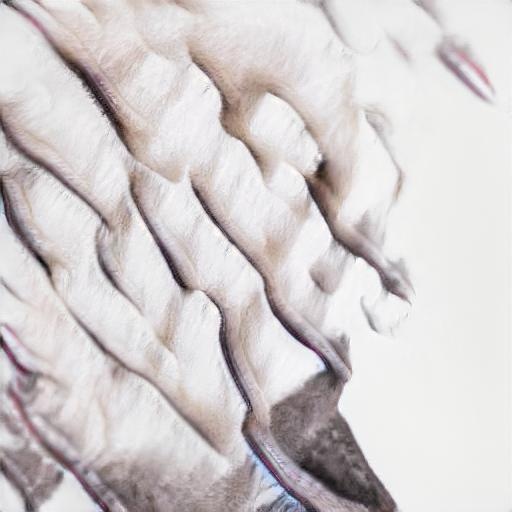} &
\includegraphics[width=0.15\linewidth]{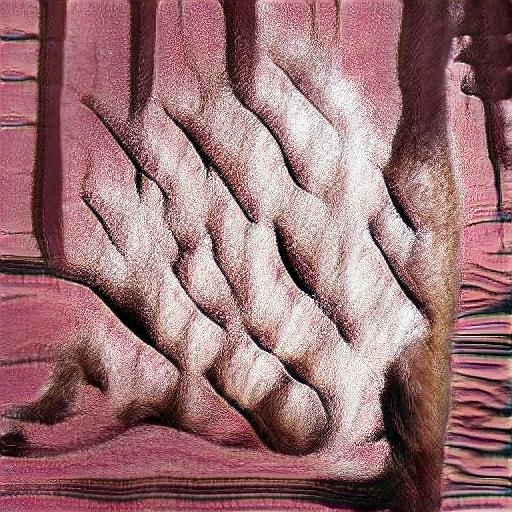} &
 \hphantom{a} &
\includegraphics[width=0.15\linewidth]{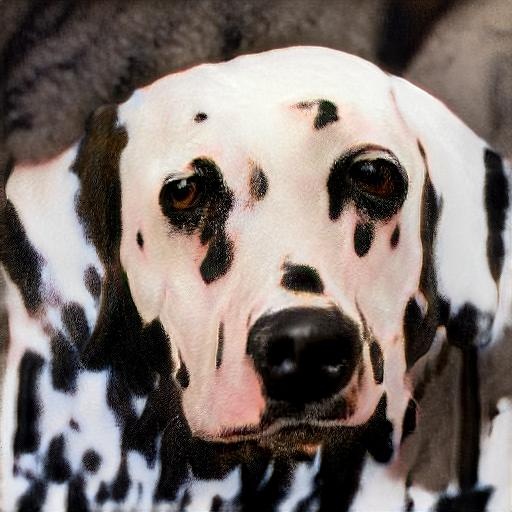} &
\includegraphics[width=0.15\linewidth]{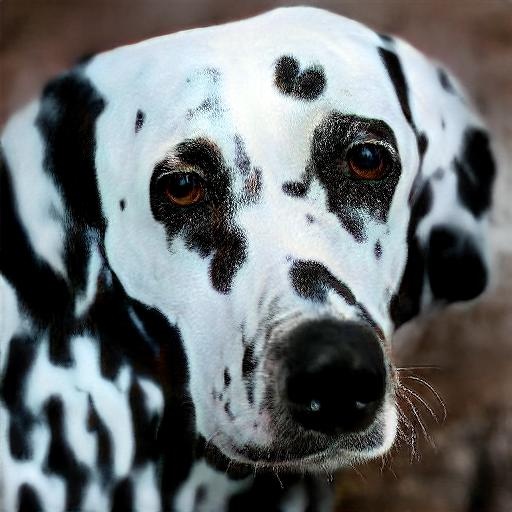} &
\includegraphics[width=0.15\linewidth]{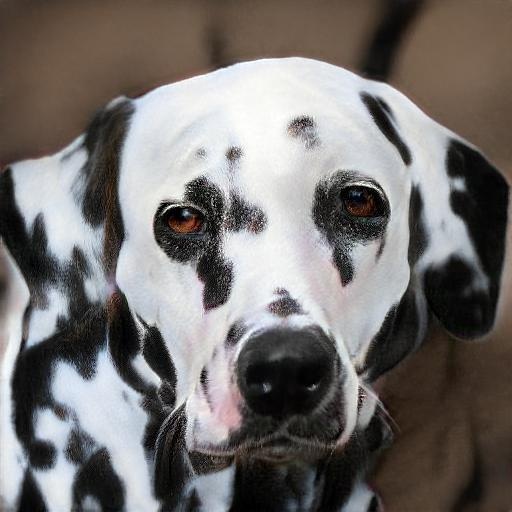} \\
  \includegraphics[width=0.15\linewidth]{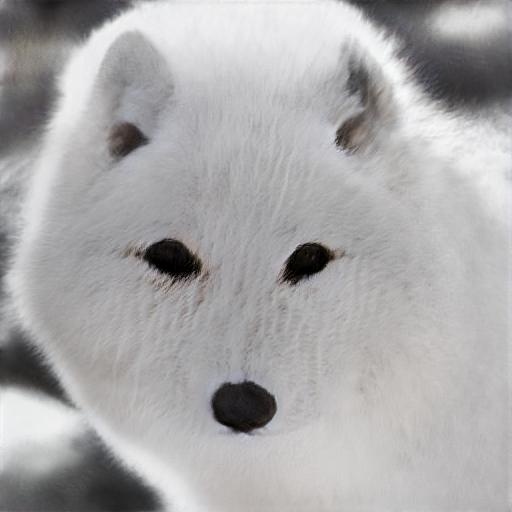} &
\includegraphics[width=0.15\linewidth]{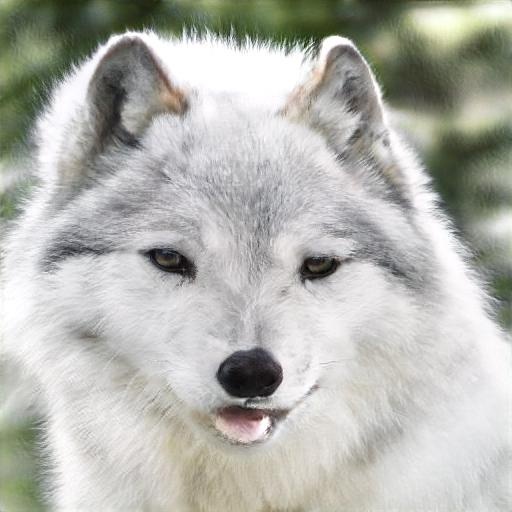} &
\includegraphics[width=0.15\linewidth]{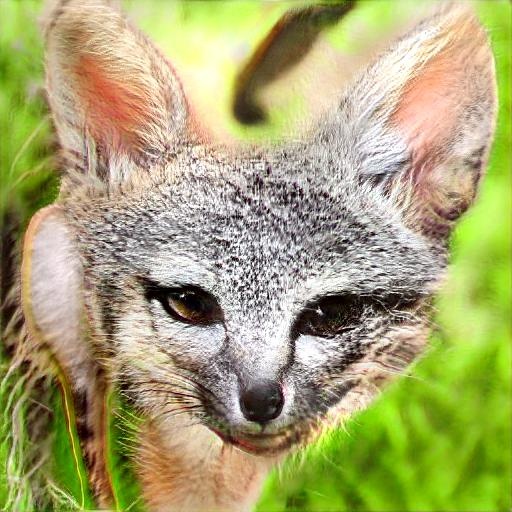} &
 \hphantom{a} &
\includegraphics[width=0.15\linewidth]{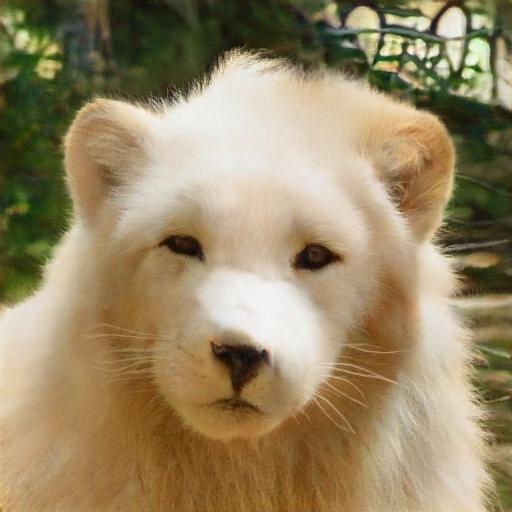} &
\includegraphics[width=0.15\linewidth]{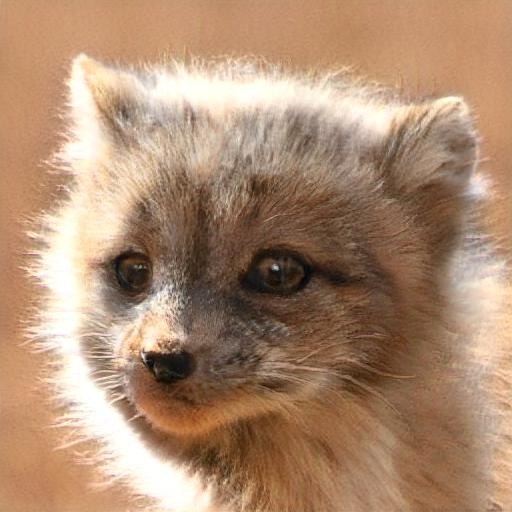} &
\includegraphics[width=0.15\linewidth]{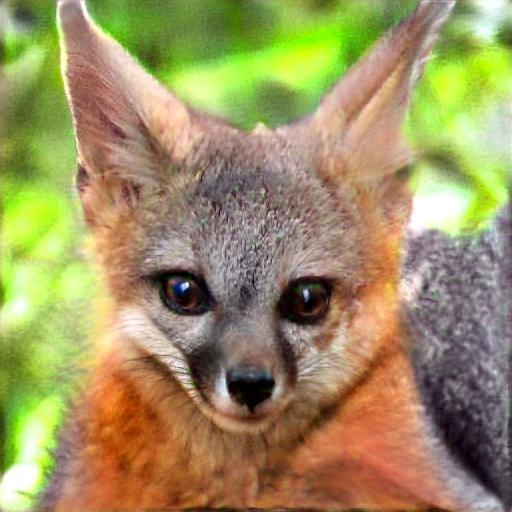} 
\end{tabular}
\caption{Worst sample comparison}\hspace{0.5cm}
\end{subfigure}
\begin{subfigure}[t]{1.0\linewidth}
\centering
\begin{tabular}[b]{ccccccccc}
 \multicolumn{4}{c}{StyleGAN2} &  \hphantom{ab} &  \multicolumn{4}{c}{StyleGAN2+GGDR}\\
\includegraphics[width=0.115\linewidth]{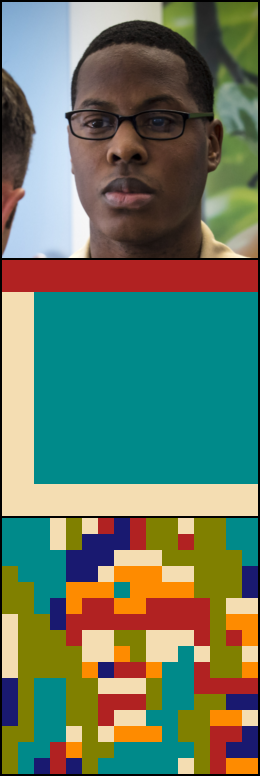} &
\includegraphics[width=0.115\linewidth]{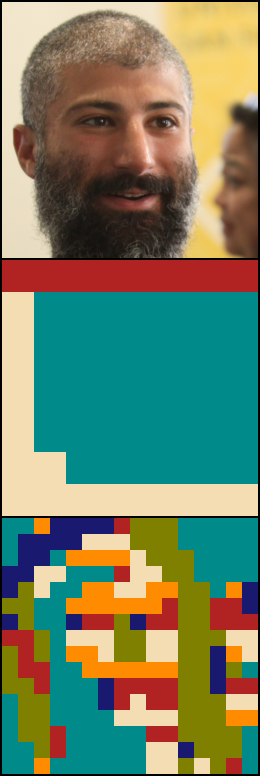} &
\includegraphics[width=0.115\linewidth]{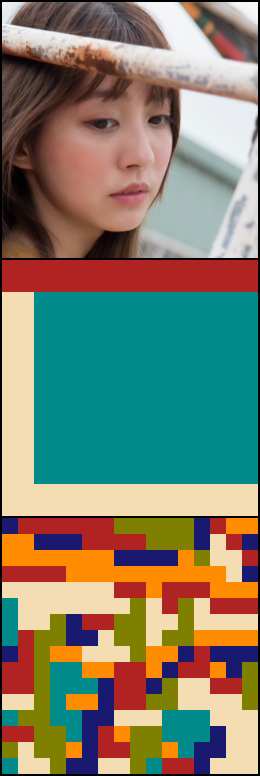} &
\includegraphics[width=0.115\linewidth]{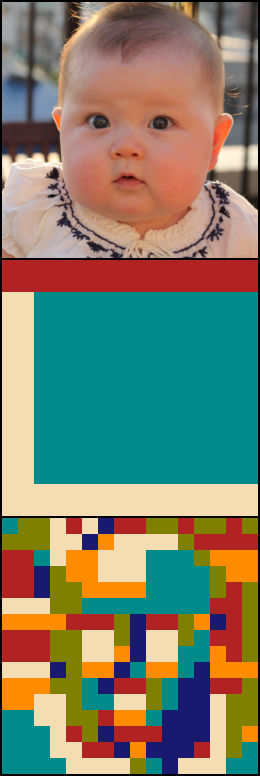} &
 \hphantom{ab} &
\includegraphics[width=0.115\linewidth]{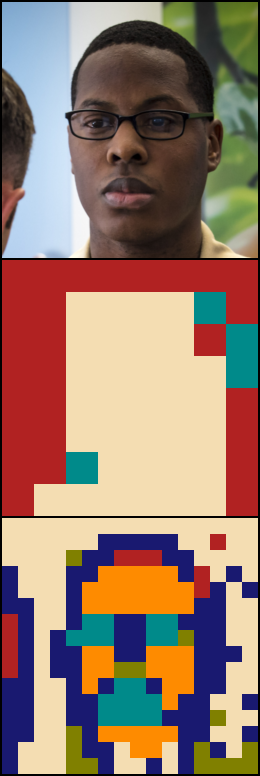} &
\includegraphics[width=0.115\linewidth]{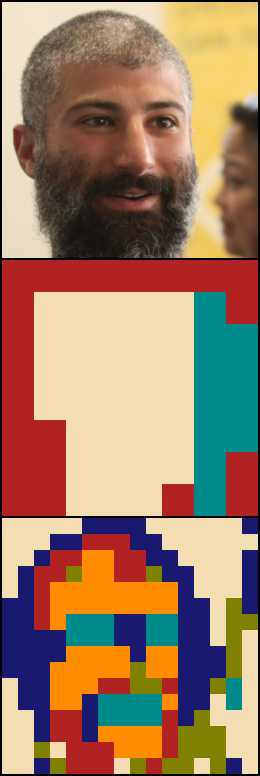} &
\includegraphics[width=0.115\linewidth]{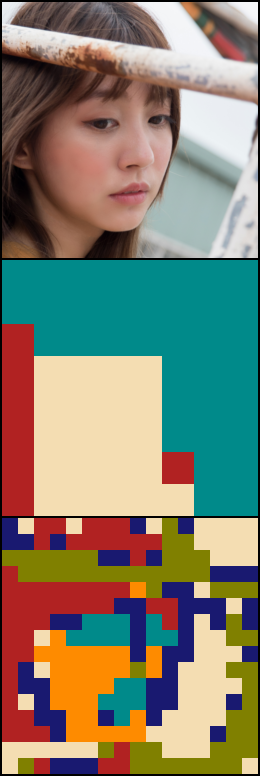} &
\includegraphics[width=0.115\linewidth]{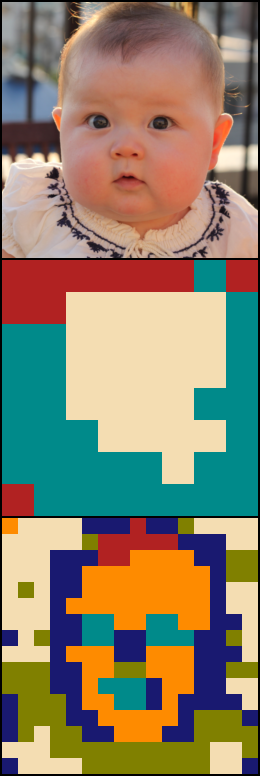}
\end{tabular}
\caption{Comparison of the the encoder feature maps in the discriminators}
\end{subfigure}
\caption{(a) Qualitative comparison of worst-sample images with and without GGDR on AFHQ. (b)$k$-means clustering of the feature maps of the encoder in the discriminator with and without GGDR loss. From top to bottom, real images, feature maps that are 8($k=3$) and 16($k=6$) pixels wide. 
}
\label{fig:worst}
\end{figure}
}

In ~\figref{comparison} and ~\figref{d_feat_vis} (a), we show some selected results of our method on the evaluated datasets. We visualize StyleGAN2 with GGDR for FFHQ and LSUN, and ADA with GGDR for other datasets. More uncurated images are shown in the supplementary material. Since our method tends to improve the recall than the precision, it is hard to show visual improvement with the limited numbers of samples. Instead, we compare the worst samples in ~\figref{worst} (a). We follow the method of ~\cite{kumari2021ensembling} to sort the samples, which uses the Inception~\cite{szegedy2016rethinking} model to fit a gaussian model and sorts by the log-likelihood using it. We can see the worst samples of our method still contain objects unlike those of ADA. It is interesting that ~\cite{kumari2021ensembling} reports similar improvements on the worst samples by utilizing pretrained models for the discriminator. We conjecture that the feature maps of the generator plays similar role with the pretrained models in their works~\cite{kumari2021ensembling, sauer2021projected}.

\subsection{Analysis and ablation study} 
\label{sec:ablexp}
\begin{figure}[t]
  \centering

  \begin{subfigure}[t]{0.7\linewidth}
        \includegraphics[width=\linewidth]{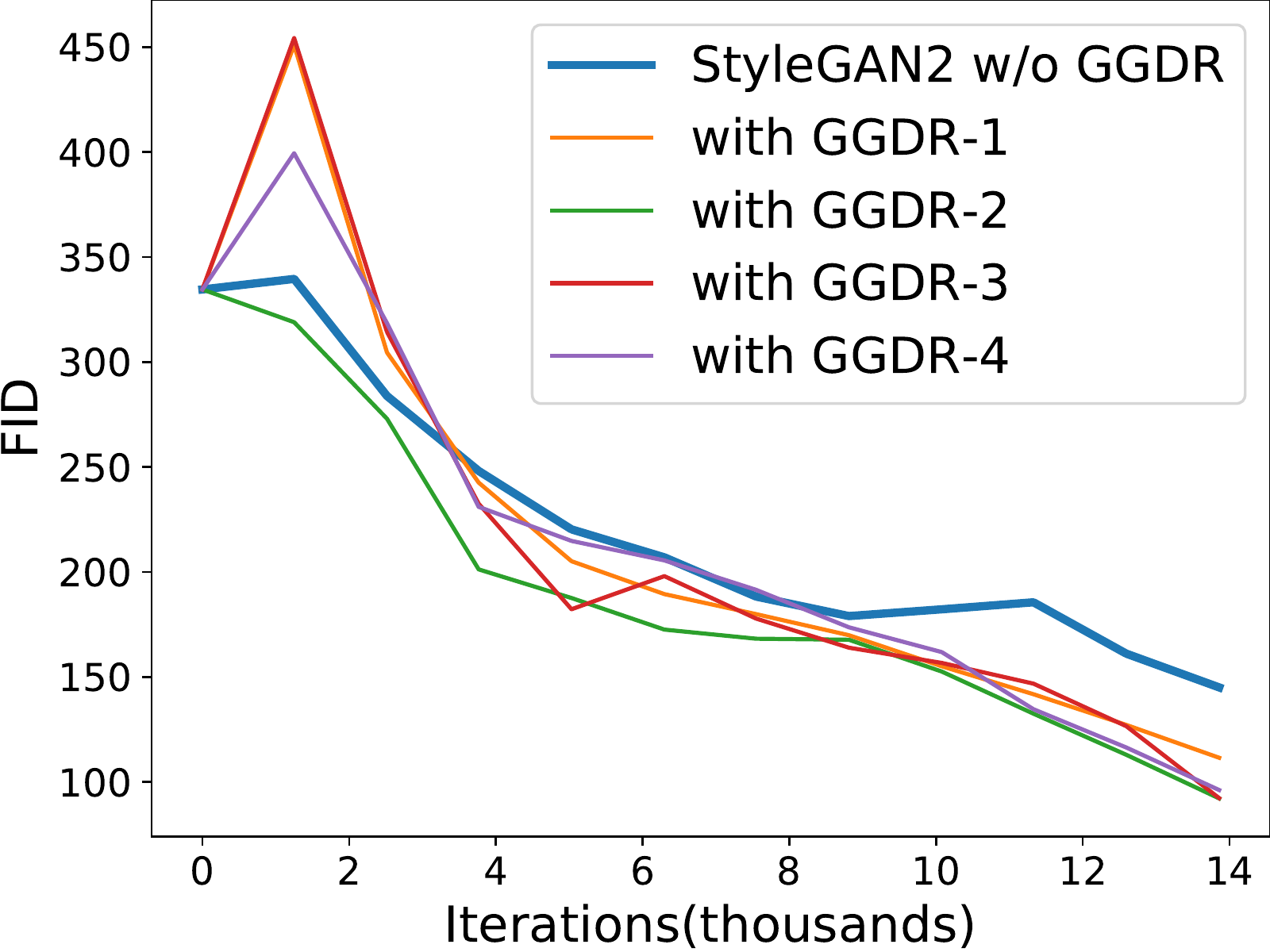}
        \caption{FIDs in the early training phase}\hspace{0.5cm}
  \end{subfigure}

\begin{subfigure}[t]{0.7\linewidth}
    \includegraphics[width=\linewidth]{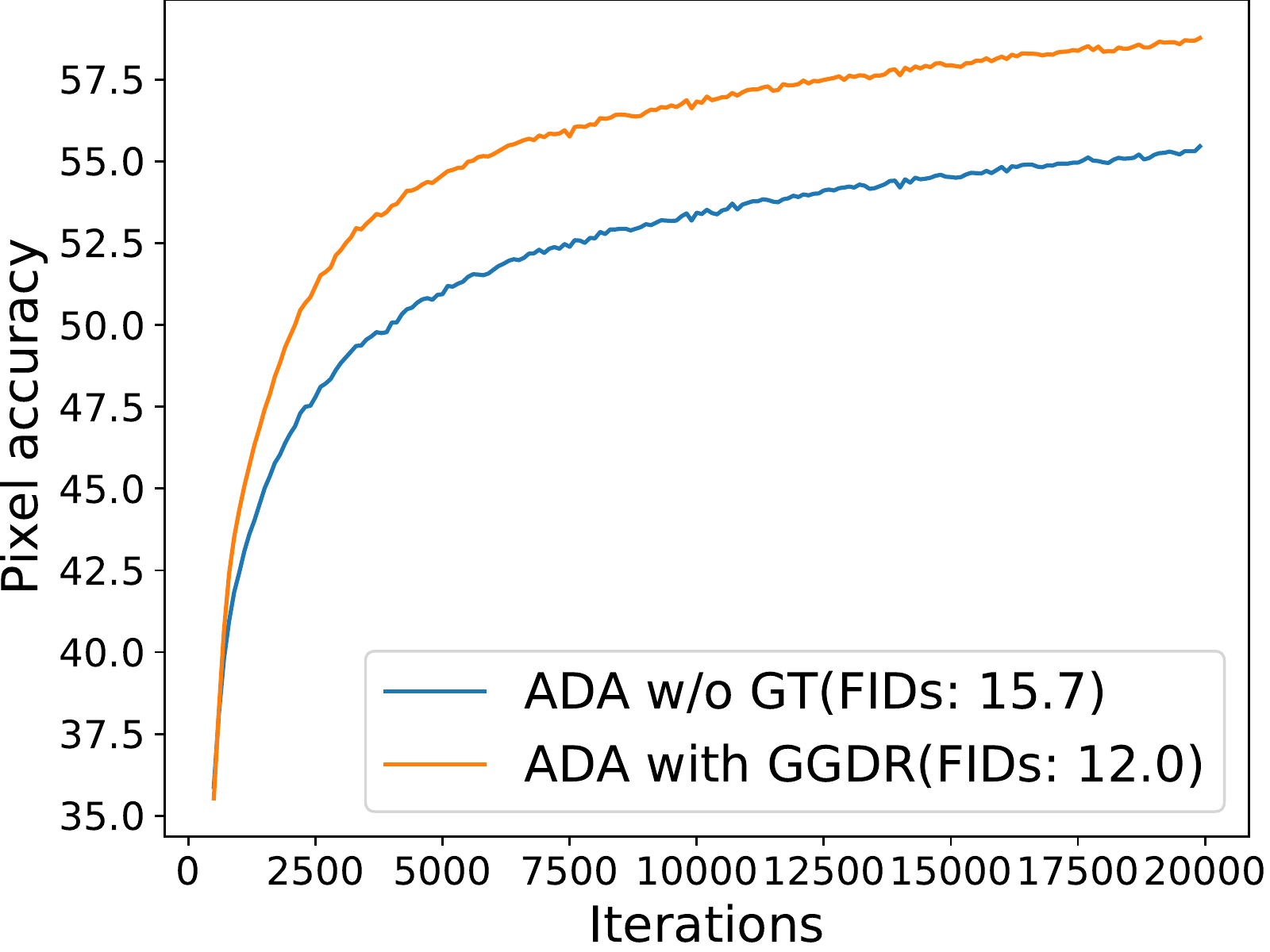}
    \caption{Downstream task result}
  \end{subfigure}
  
  \caption{(a) FID score graph of multiple experiments with GGDR in the early training phase on LSUN Cat. (b) Validation pixel accuracy on the ADE20K segmentation task with the frozen discriminator trained with and without GGDR. 
  }
  \label{fig:downstream}
\end{figure}

The proposed GGDR supervises a discriminator using the intermediate feature maps of a generator in a GAN model, so it depends on the quality of the generator feature maps. In \secref{analysis_feature_maps}, we show that the feature maps of the generator contain valid semantic information even in the early stage. In addition to the visualization, we measure the FIDs in early iterations and show in \figref{downstream} (a). To check the effect of initializations, we run multiple experiments on LSUN Cat. In early iterations, GGDR can interfere the performance by using less trained feature maps and bad initialization. However, after only several thousand iterations, StyleGAN2 with GGDR starts to converge faster and shows better scores. 
{
\setlength{\tabcolsep}{0pt}
\renewcommand{\arraystretch}{0.7}
\begin{figure}[t]
  \centering
\begin{subfigure}[t]{0.45\linewidth}
    \centering
        \includegraphics[width=\linewidth]{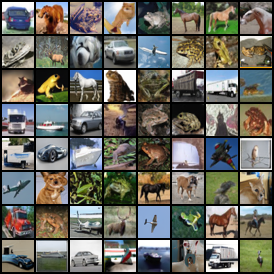}
\caption{Samples on CIFAR10}
  \end{subfigure}\hfill
\begin{subfigure}[t]{0.45\linewidth}
    \centering
        \includegraphics[width=\linewidth]{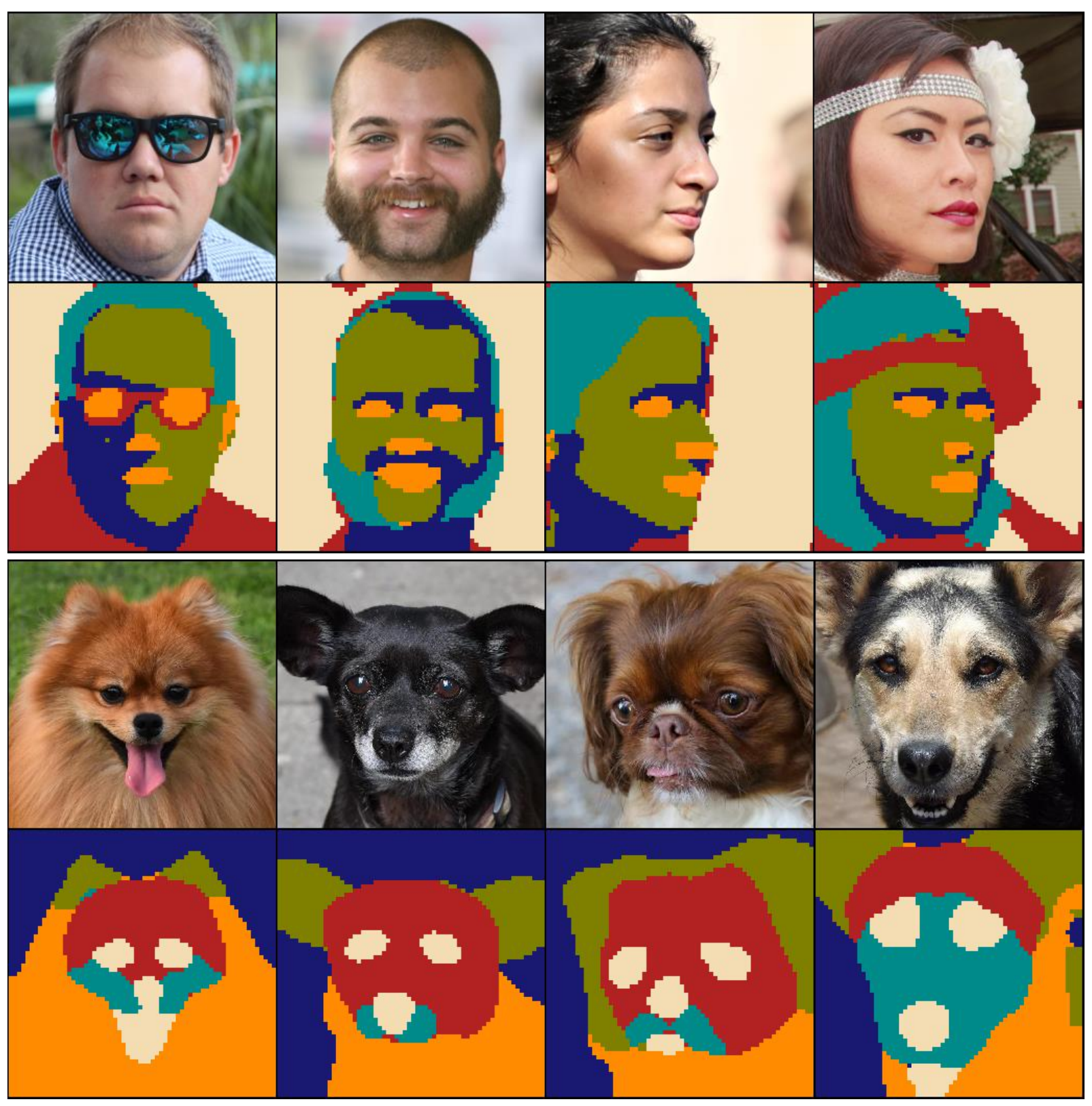}
        \caption{Visualization of the decoder output on real samples}
  \end{subfigure}
    \caption{(a) Random samples by ADA with GGDR on CIFAR-10 dataset. (b) $k$-means clustering of the feature maps of the decoder in our discriminator on real images($k=6$).
  }
  \label{fig:d_feat_vis}
\end{figure}
}

\begin{table}[t]
  \centering
   \begin{subtable}[b]{0.45\linewidth}
     \centering
  \begin{tabular}{p{2cm} c}
\toprule
Target & FID \\
\midrule
None & 7.98 \\
$8\times8$ & 7.57 \\
$16\times16$ & 6.56 \\
$32\times32$ & 5.98 \\
$64\times64$ & \textbf{5.28} \\ 
\bottomrule
\end{tabular}
\caption{Target size}\hspace{0.5cm}
 \end{subtable}\hfill
 \begin{subtable}[b]{0.45\linewidth}
 \centering
 {
  \begin{tabular}{p{2cm} c}
\toprule
Activation & FID \\
\midrule
Linear & \textbf{5.28} \\
leaky ReLU & 5.43 \\
\bottomrule
\toprule
Kernel size & FID \\
\midrule
$1\times1$ & 5.28 \\
$3\times3$ & \textbf{5.25} \\ 
\bottomrule
\end{tabular}\hspace{0.5cm}
}
\caption{Decoder design}\hspace{0.5cm}
  \end{subtable}
\begin{subtable}[b]{1.0\linewidth}
\centering
{
\begin{tabular}{c l l }
\toprule
Method & \# params & time(s)\\ 
\midrule
Baseline & 4.87M & 5.60\\
+ GGDR & 5.05M & 6.05\\
\midrule
& (+3.7\%) & (+8.0\%) \\
\bottomrule
\end{tabular}
}
\caption{Calculation costs}\label{tab:1b}
\end{subtable}%
  \caption{Ablation studies and calculation costs on LSUN Cat with eight V100 GPUs. Ablation study on (a) the target feature map size and (b) the decoder design. (c) Calculation costs with and without GGDR.}
  \label{tab:ablation}
\end{table}

In \tabref{ablation}, we conduct ablation studies to investigate the effects of the decoder architecture and the feature map resolution. In our experiments, we select $64 \times 64$ feature maps as the guidance. One may curious the performance differences if we use different sizes of the guidance feature maps. As shown in \tabref{ablation} (a), utilizing the large and dense feature maps achieves the best FID scores. Meanwhile, we design a compact decoder with $1 \times 1$ convolutional filters and linear activation for fast training and convergence. In \tabref{ablation} (b), we show that changing decoder activation and kernel size affects only negligible performance difference. 
{
\setlength{\tabcolsep}{1pt}
\renewcommand{\arraystretch}{0}
\begin{figure*}[t]
\begin{center}
  \begin{tabular}{cccccccc}
FFHQ & \makecell{LSUN\\Cat} & \makecell{LSUN\\Horse} & \makecell{LSUN\\Church} & \makecell{AFHQ\\Cat} & \makecell{AFHQ\\Dog} & \makecell{AFHQ\\Wild} & \makecell{Lands-\\scape} \\
\includegraphics[width=0.118\linewidth]{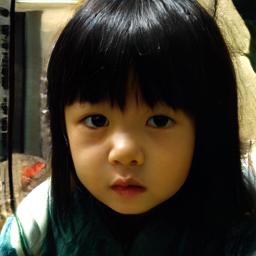} &
\includegraphics[width=0.118\linewidth]{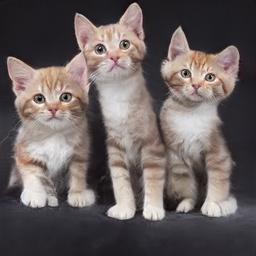} &
\includegraphics[width=0.118\linewidth]{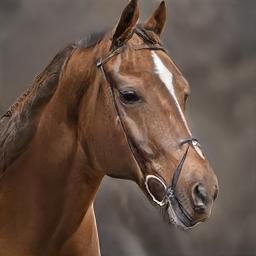} &
\includegraphics[width=0.118\linewidth]{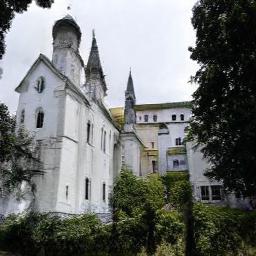} &
\includegraphics[width=0.118\linewidth]{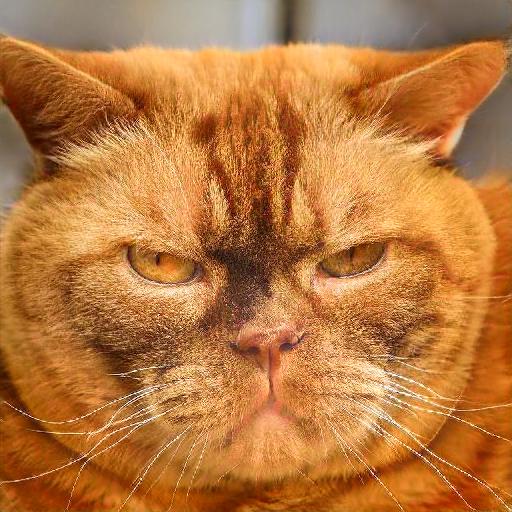} &
\includegraphics[width=0.118\linewidth]{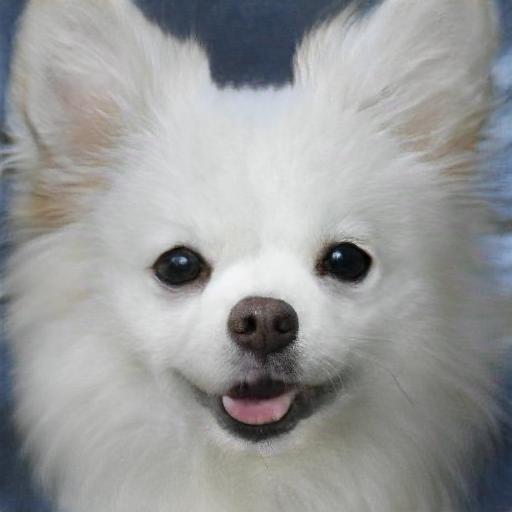} &
\includegraphics[width=0.118\linewidth]{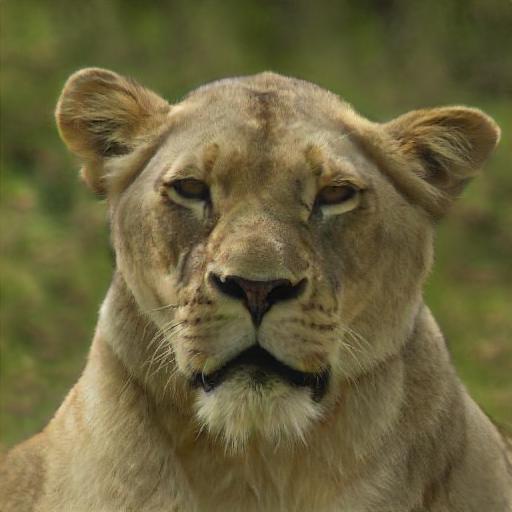} &
\includegraphics[width=0.118\linewidth]{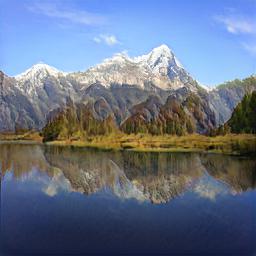} \\

\includegraphics[width=0.118\linewidth]{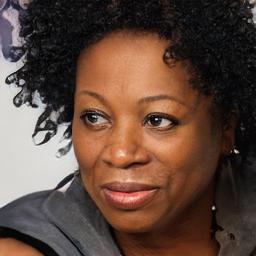} &
\includegraphics[width=0.118\linewidth]{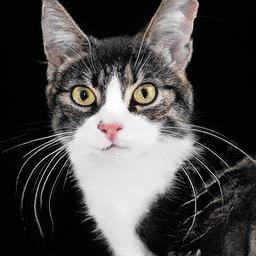} &
\includegraphics[width=0.118\linewidth]{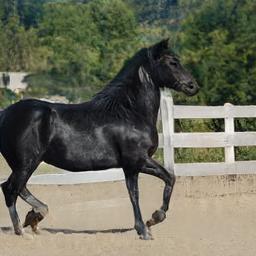} &
\includegraphics[width=0.118\linewidth]{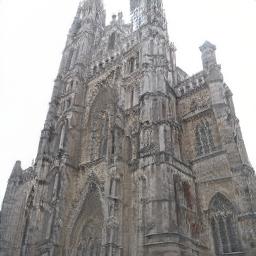} &
\includegraphics[width=0.118\linewidth]{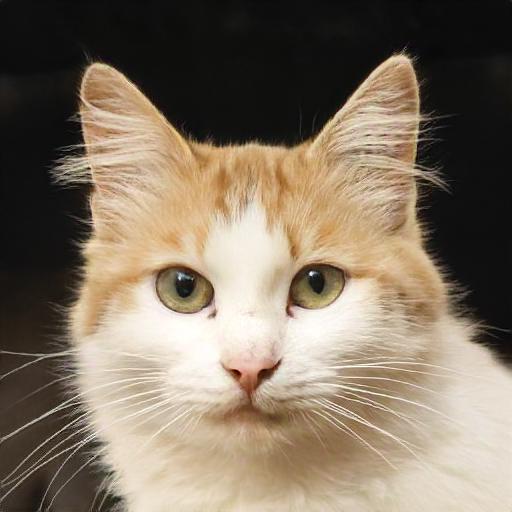} & 
\includegraphics[width=0.118\linewidth]{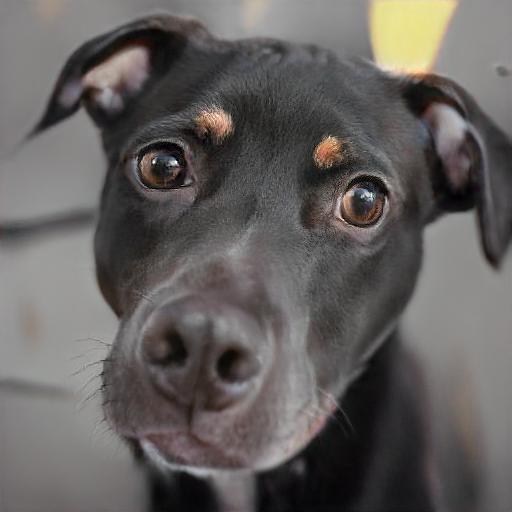} &
\includegraphics[width=0.118\linewidth]{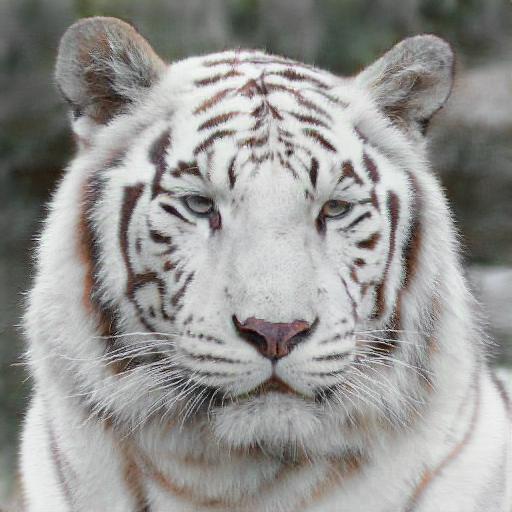} &
\includegraphics[width=0.118\linewidth]{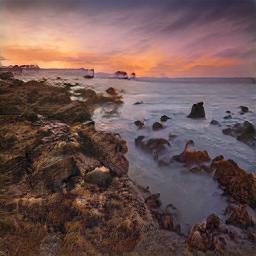} \\

\includegraphics[width=0.118\linewidth]{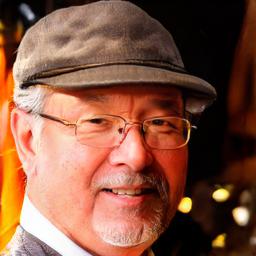} &
\includegraphics[width=0.118\linewidth]{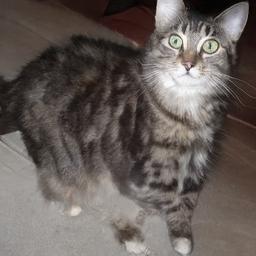} &
\includegraphics[width=0.118\linewidth]{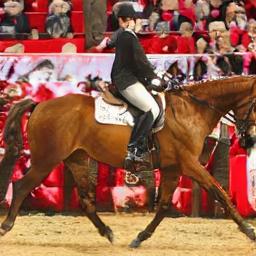} &
\includegraphics[width=0.118\linewidth]{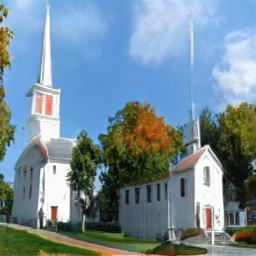} &
\includegraphics[width=0.118\linewidth]{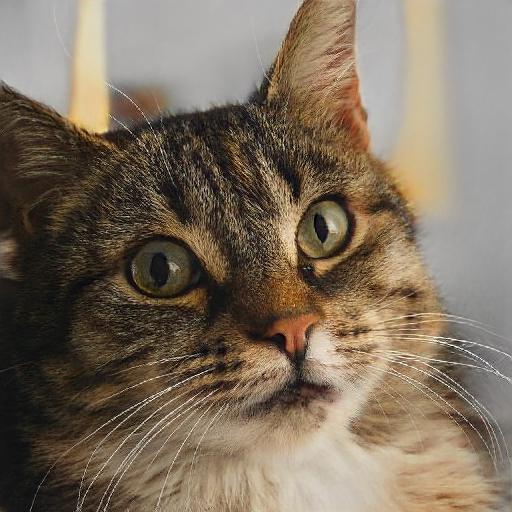} & 
\includegraphics[width=0.118\linewidth]{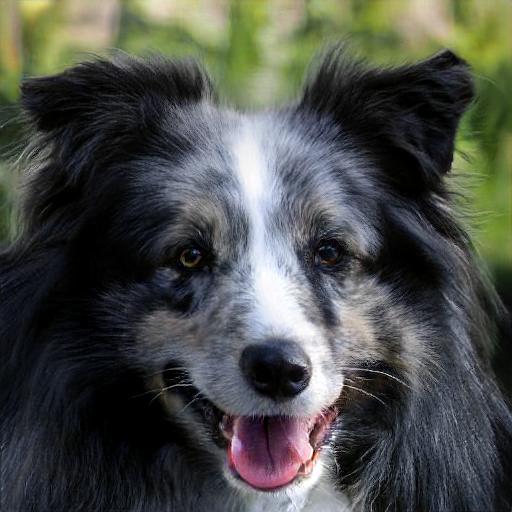} &
\includegraphics[width=0.118\linewidth]{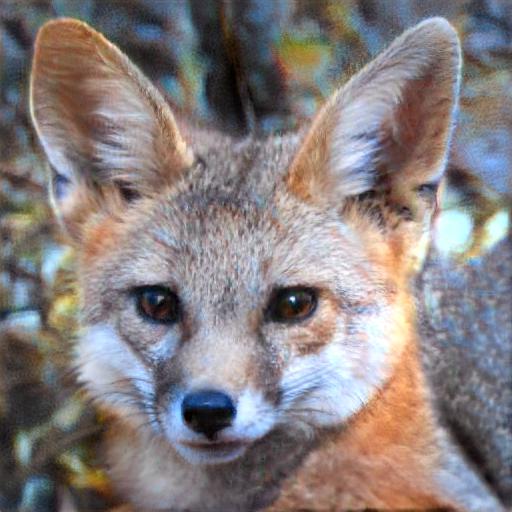} &
\includegraphics[width=0.118\linewidth]{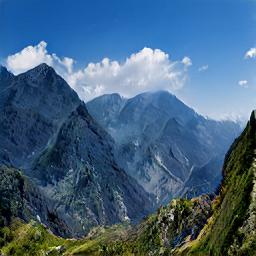} \\

\includegraphics[width=0.118\linewidth]{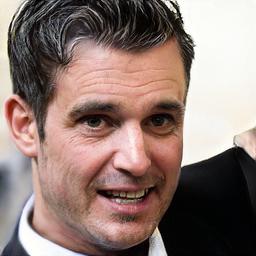} &
\includegraphics[width=0.118\linewidth]{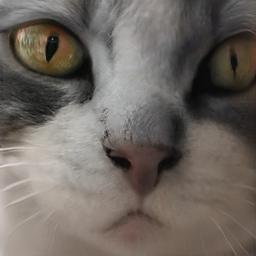} &
\includegraphics[width=0.118\linewidth]{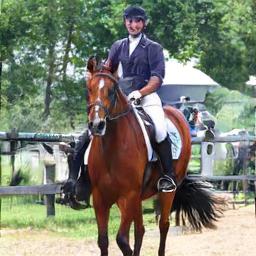} &
\includegraphics[width=0.118\linewidth]{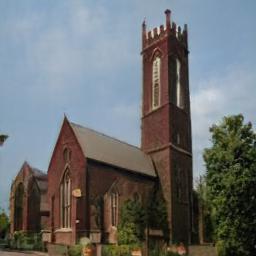} &
\includegraphics[width=0.118\linewidth]{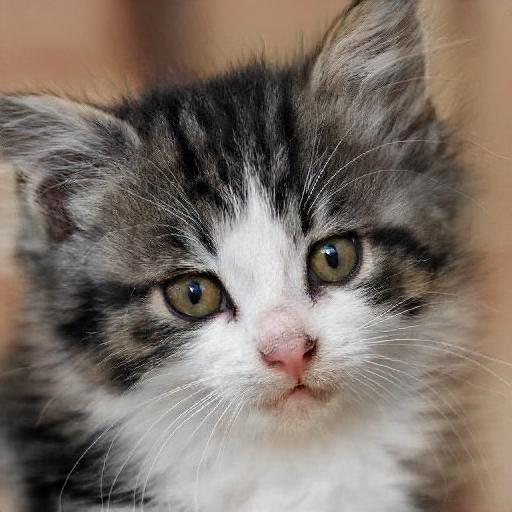} &
\includegraphics[width=0.118\linewidth]{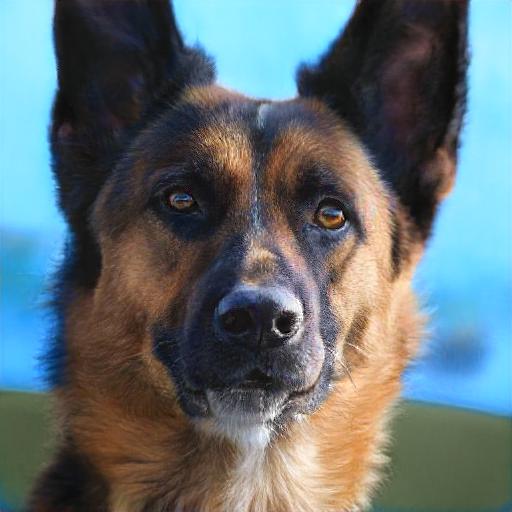} &
\includegraphics[width=0.118\linewidth]{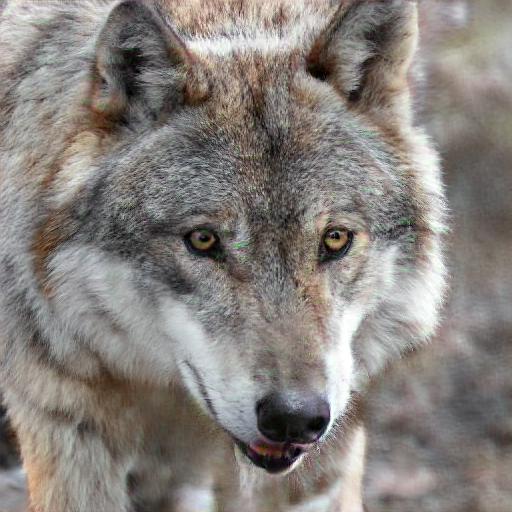} &
\includegraphics[width=0.118\linewidth]{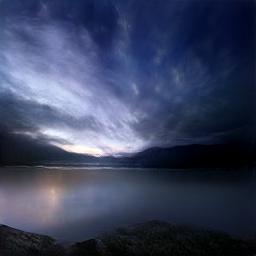} \\

\includegraphics[width=0.118\linewidth]{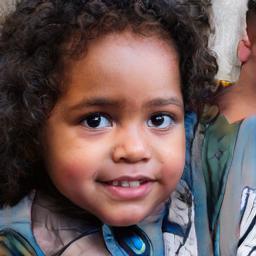} &
\includegraphics[width=0.118\linewidth]{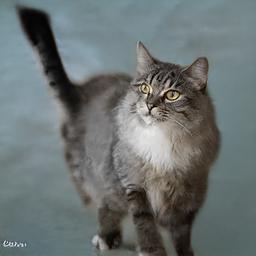} &
\includegraphics[width=0.118\linewidth]{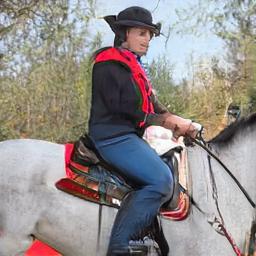} &
\includegraphics[width=0.118\linewidth]{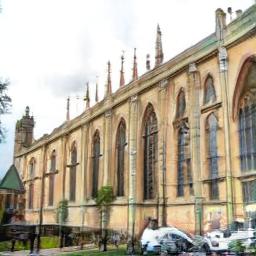} &
\includegraphics[width=0.118\linewidth]{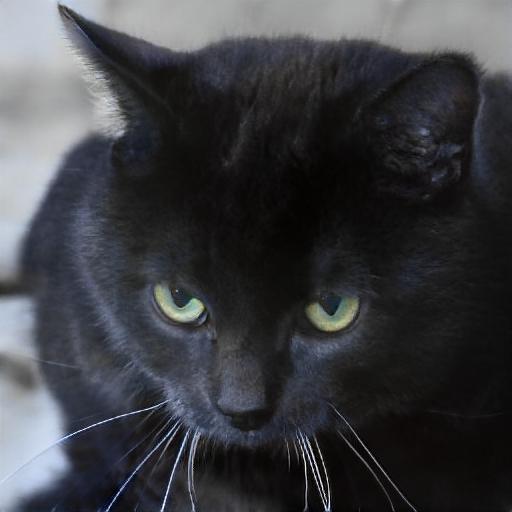} &
\includegraphics[width=0.118\linewidth]{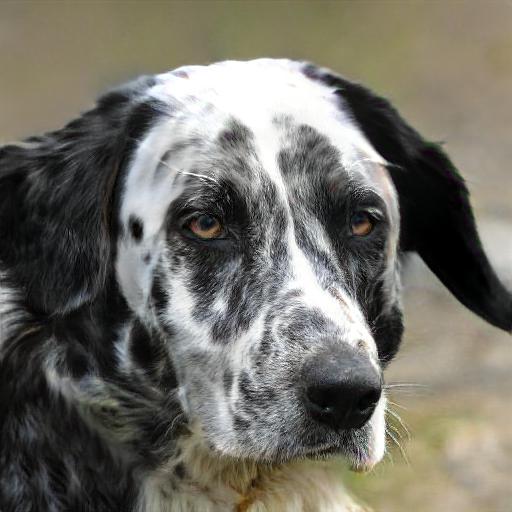} &
\includegraphics[width=0.118\linewidth]{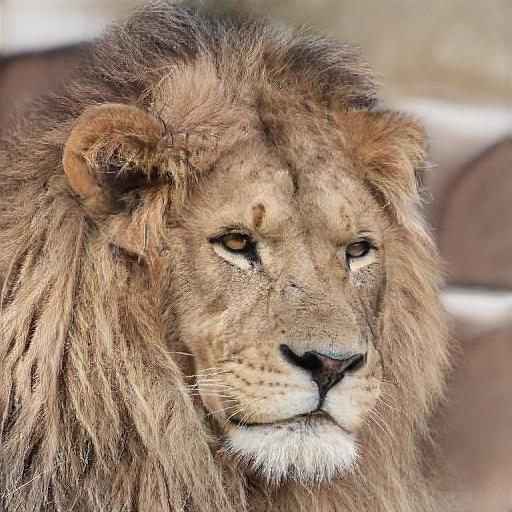} &
\includegraphics[width=0.118\linewidth]{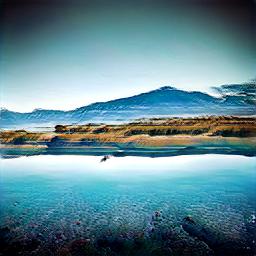} \\
\end{tabular}
\end{center}
   \caption{Selective samples generated by our method. For FFHQ and LSUN datasets, we show the results of StyleGAN2 with GGDR. For AFHQ and Landscape, we show the results of ADA with GGDR.}
\label{fig:comparison}
\end{figure*}
}

To analyze the effects of GGDR, we visualize and compare the original discriminator part in \figref{worst} (b). We run $k$-means clustering as done in \secref{analysis_feature_maps}. With GGDR loss, the shared encoder part prefers to learn high-level features which are useful for both tasks, so its feature maps reveal more semantically meaningful clusters. By guiding to learn semantic features, our approach can help discriminators to focus on salient parts of the image instead of meaningless features. To further analyze, we conduct a downstream experiment, training a shallow segmentation network using the extracted features by the discriminator with or without GGDR. As shown in \figref{downstream} (b), the accuracy on validation data shows the discriminator with GGDR has more representation power on semantic information. Meanwhile, since we trained on fake images only, it may be curious if the guidance by fake images is still valid for real images. In \figref{d_feat_vis} (b), we visualize the outputs of the decoder of our discriminator on real image. While the decoder of our method learned using fake images, we can see that its features well capture the semantically meaningful regions of the real images.

In \tabref{ablation} (c), we show the additional calculation costs when use GGDR. We can see the additional costs are marginal where the parameters increase $3.7\%$ and the time increases $8.0\%$. For these measurement, we run the StyleGAN2 on the $256 \times 256$ dataset with eight V100 GPUs.

\section{Related Work}
\label{sec:related_work}

\textbf{Conditional image synthesis utilizing semantic label map.}
For the controllability of the generated images, it is common to exploit semantic layout-level information for the conditional image generation~\cite{isola2017pix2pix,wang2018pix2pixHD,liu2019ccfpse}. SPADE~\cite{park2019spade} utilizes Spatilally-Adaptive Denormalization which preserves semantic informations, and OASIS~\cite{sushko2020oasis} have shown that it is able to train the conditional GANs using the discriminator that predicts pixel-level semantic labels, without incorporating semantic maps as additional conditions. Also, instead of using explicit semantic ground-truths, it is possible to use the features from the deep networks for the semantic guidance of the generator as shown in~\cite{shocher2020semantic,ditria2020metric}. Unlike these works, our method aims at unconditional image generation.

\textbf{Regularization for GANs.}
Several works interest to stabilizing the GAN trainings, especially by regularizing the discriminators~\cite{kim2021feature, miyato2018spectral, mescheder2018r1reg}. Recently, utilizing augmentations for the discriminator gained a lot of interests, which was proven successful in general vision tasks~\cite{yun2019cutmix,zhang2017mixup}. In consistency regularization (CR)~\cite{zhang2019consistency,zhao2020improved}, in addition to using regular GAN losses, a discriminator is penalized by the differences in the outputs between augmented and non-augmented images. APA~\cite{jiang2021apa} regularizes the discriminator by utilizing fake images as psuedo-real data adaptively. DiffAugment~\cite{zhao2020differentiable} and ADA~\cite{karras2020ada} use non-leaking augmentations for both generator and discriminator losses. Recently, several papers use pretrained models to help discriminator for fast and stable training. ProjGAN~\cite{sauer2021projected} uses EfficientNet~\cite{tan2019efficientnet} as a feature extractor for the discriminator, and Vision-aided GAN ~\cite{kumari2021ensembling} provides automatic selection from model bank of pretrained networks to get optimal features for real and fake discrimination.

\textbf{Utilization of generator features.}
Recent studies have shown that the generators contain rich and disentangled semantic structures in the features. Collins et al.~\cite{collins2020editing} show that by applying $k$-means clustering on feature activations of the generator is possible to extract semantic objects and object parts in the generated images. Xu et al.~\cite{xu2021linear} have trained linear mapping between feature maps in the generator and semantic maps, and Endo et al.~\cite{endo2021fewshotsmis} used the nearest neighbor matching between feature maps and representative vectors by averaging the feature vectors corresponding to the ground-truth semantic labels of inverted images. StyleMapGAN~\cite{kim2021exploiting} has used spatial dimensions in the latent codes and grouping of the channels to further disentangle spatial semantic features.

\section{Conclusion and limitation}
In this paper, we present the efficacy of the dense semantic label maps for unconditional image generation. Inspired by this observation, we propose a new regularization method to leverage the feature maps of the generator instead of human annotating ground-truth semantic annotations to allow the discriminator to learn richer semantic representation. With negligible additional parameters and no ground-truth semantic segmentation map, the proposed GGDR consistently outperforms strong baselines. Since our method depends on the performance of the generator, if the generator cannot learn meaningful representations due to the extremely limited number of data or initial training collapse, GGDR will fail to improve the performance. However, thanks to modern GANs and training techniques, we believe our method can be easily applied in various situations.

\subsubsection*{Acknowledgement.} The experiments in the paper were conducted on NAVER Smart Machine Learning (NSML) platform~\cite{sung2017nsml,kim2018nsml}. We thank to Jun-Yan Zhu, Jaehoon Yoo, NAVER AI LAB researchers and the reviewers for their helpful comments and discussion.

{\small
\bibliographystyle{ieee_fullname}
\bibliography{egbib}
}

\appendix
\section{Additional results and visualization}
\begin{figure*}[t]
\centering
\begin{subfigure}{\linewidth}
\centering
\includegraphics[width=0.75\textwidth]{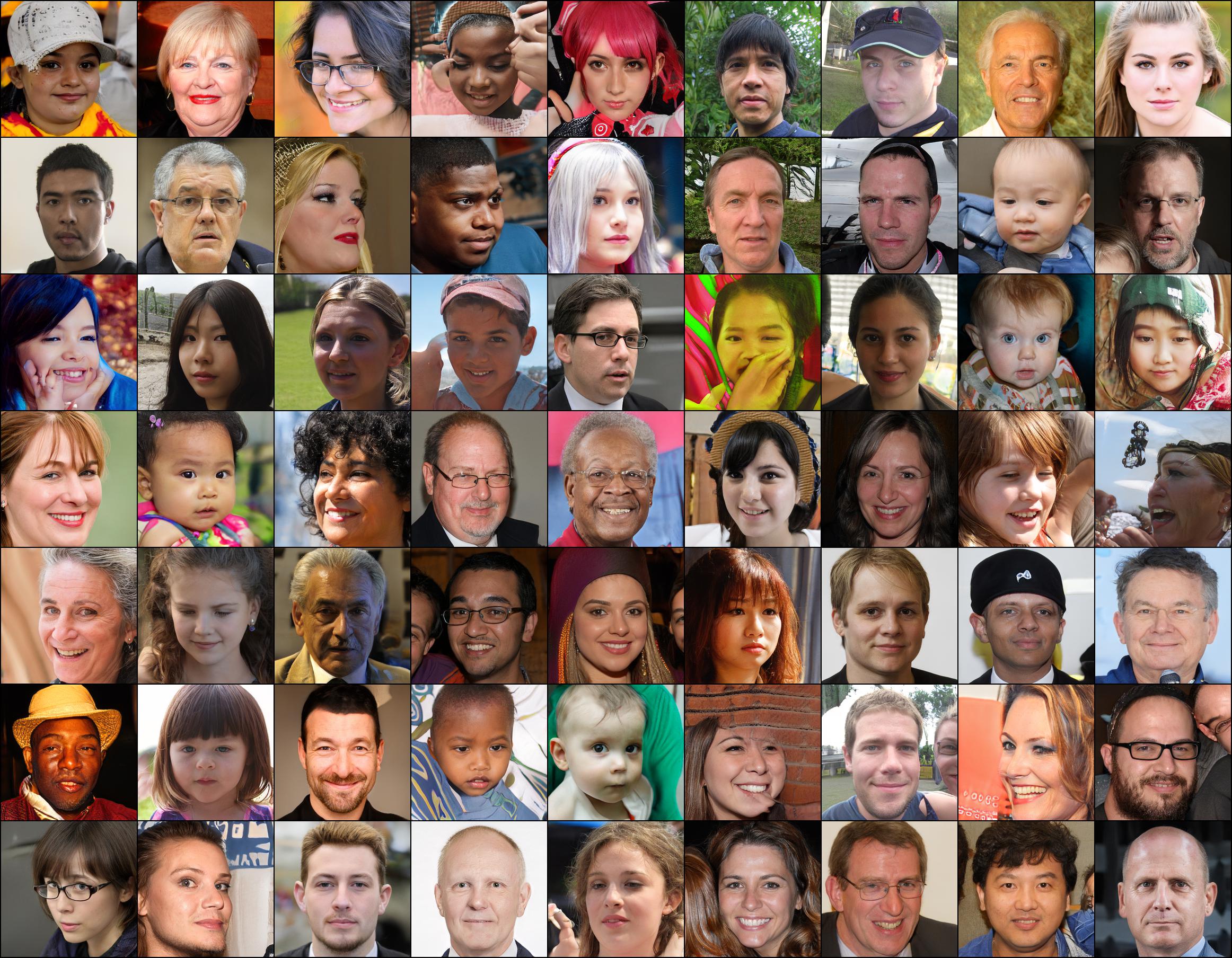}
\end{subfigure}
\par\bigskip
\begin{subfigure}{\linewidth}
\centering
\includegraphics[width=0.75\textwidth]{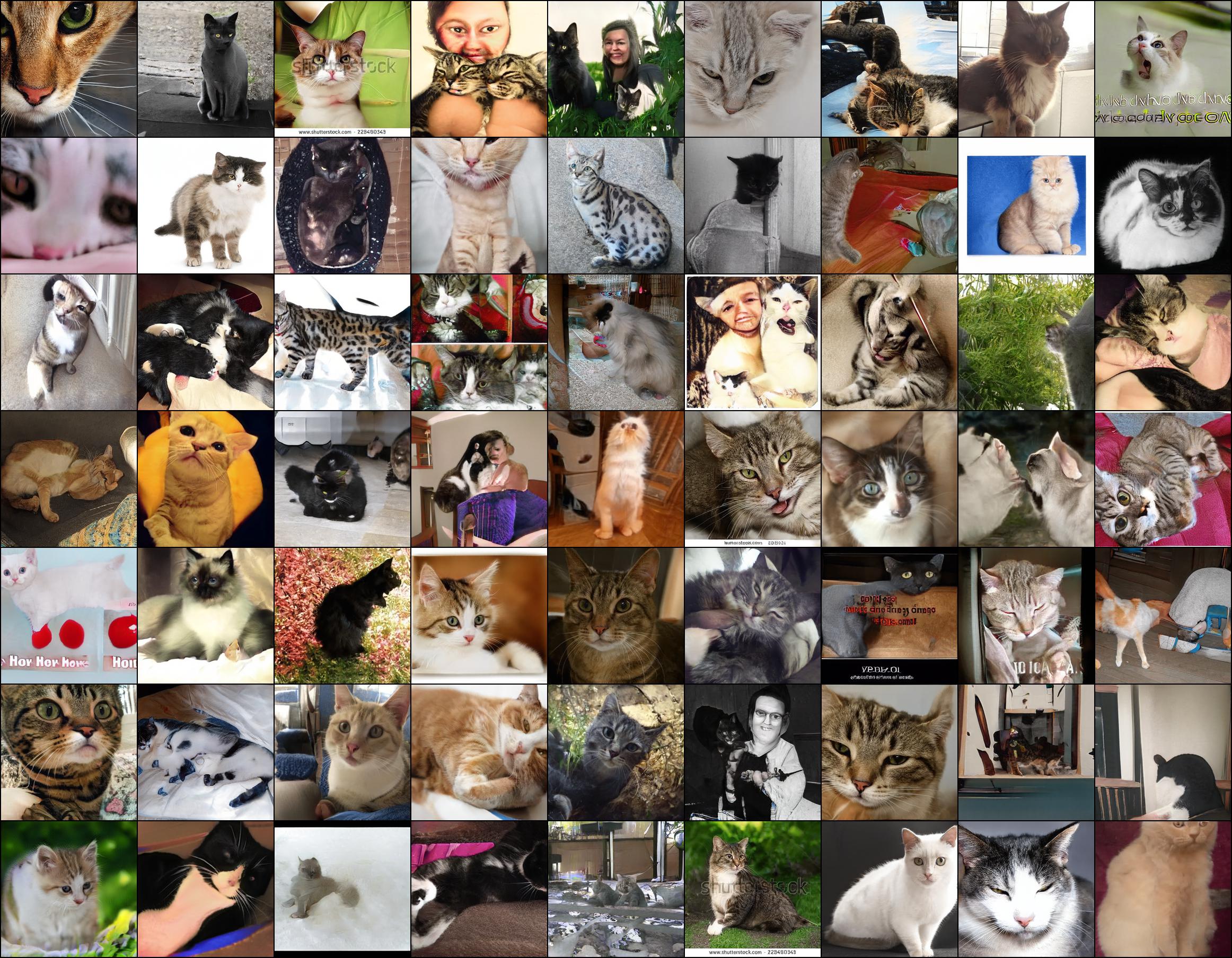}
\end{subfigure}
\caption{Uncurated results on (top) FFHQ and (bottom) LSUN Cat.} 
\label{fig:vis_ffhq_lsuncat}
\end{figure*}

\begin{figure*}[t]
\centering
\begin{subfigure}{\linewidth}
\centering
\includegraphics[width=0.75\textwidth]{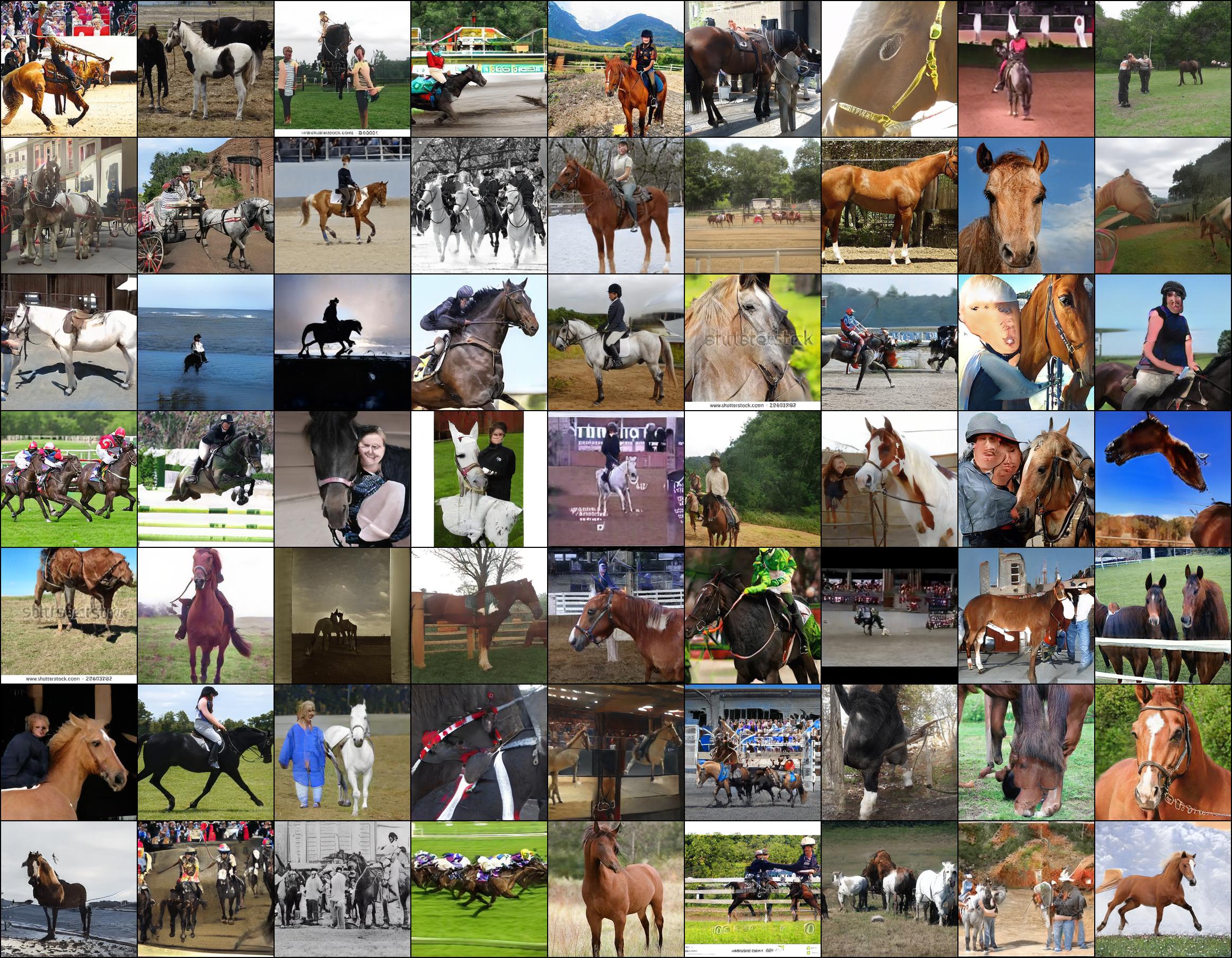}
\end{subfigure}
\par\bigskip
\begin{subfigure}{\linewidth}
\centering
\includegraphics[width=0.75\textwidth]{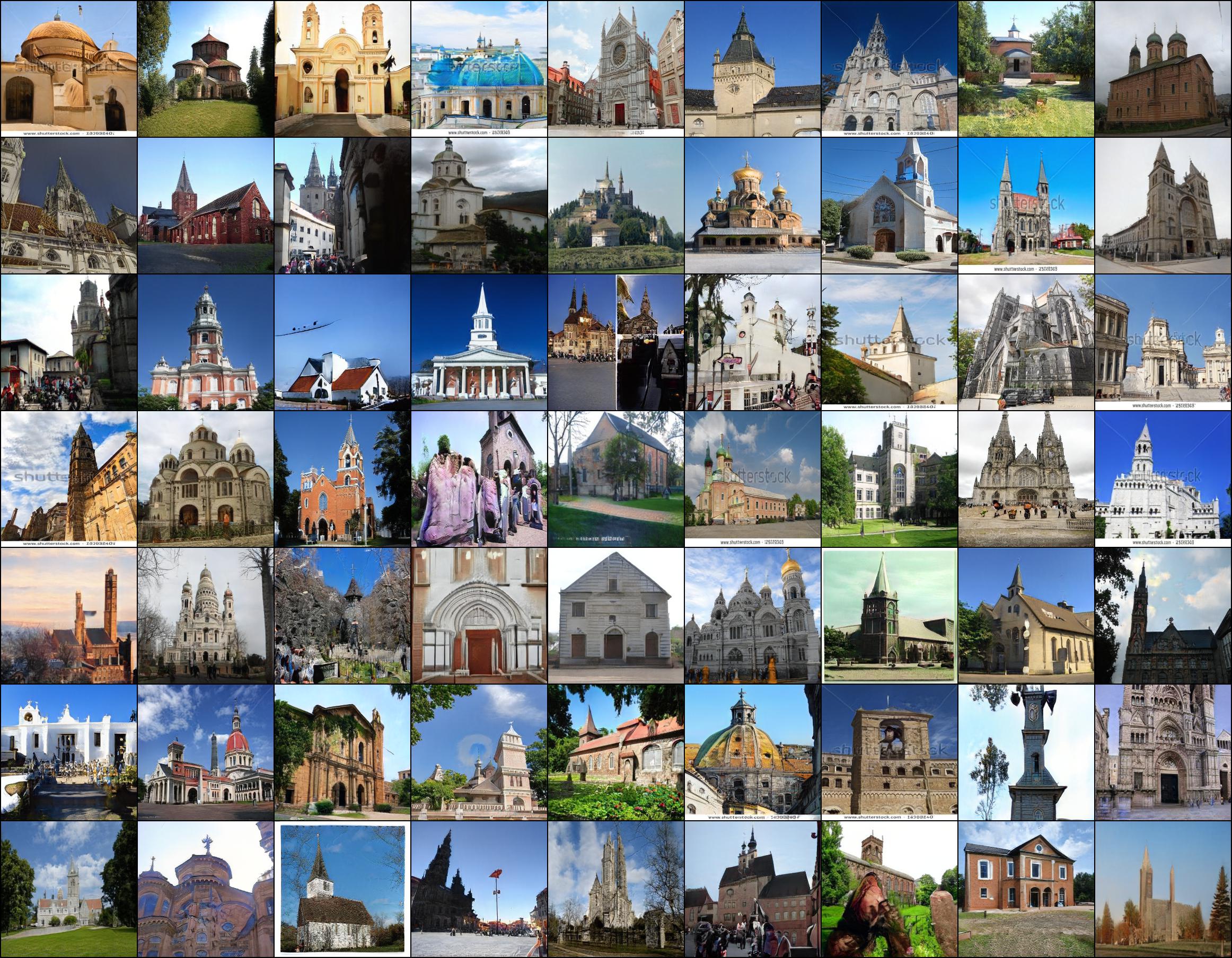}
\end{subfigure}
\caption{Uncurated results on (top) LSUN Horse and (bottom) LSUN Church.} 
\label{fig:vis_lsunhorse_church}
\end{figure*}

\begin{figure*}[t]
\centering
\begin{subfigure}{\linewidth}
\centering
\includegraphics[width=0.75\textwidth]{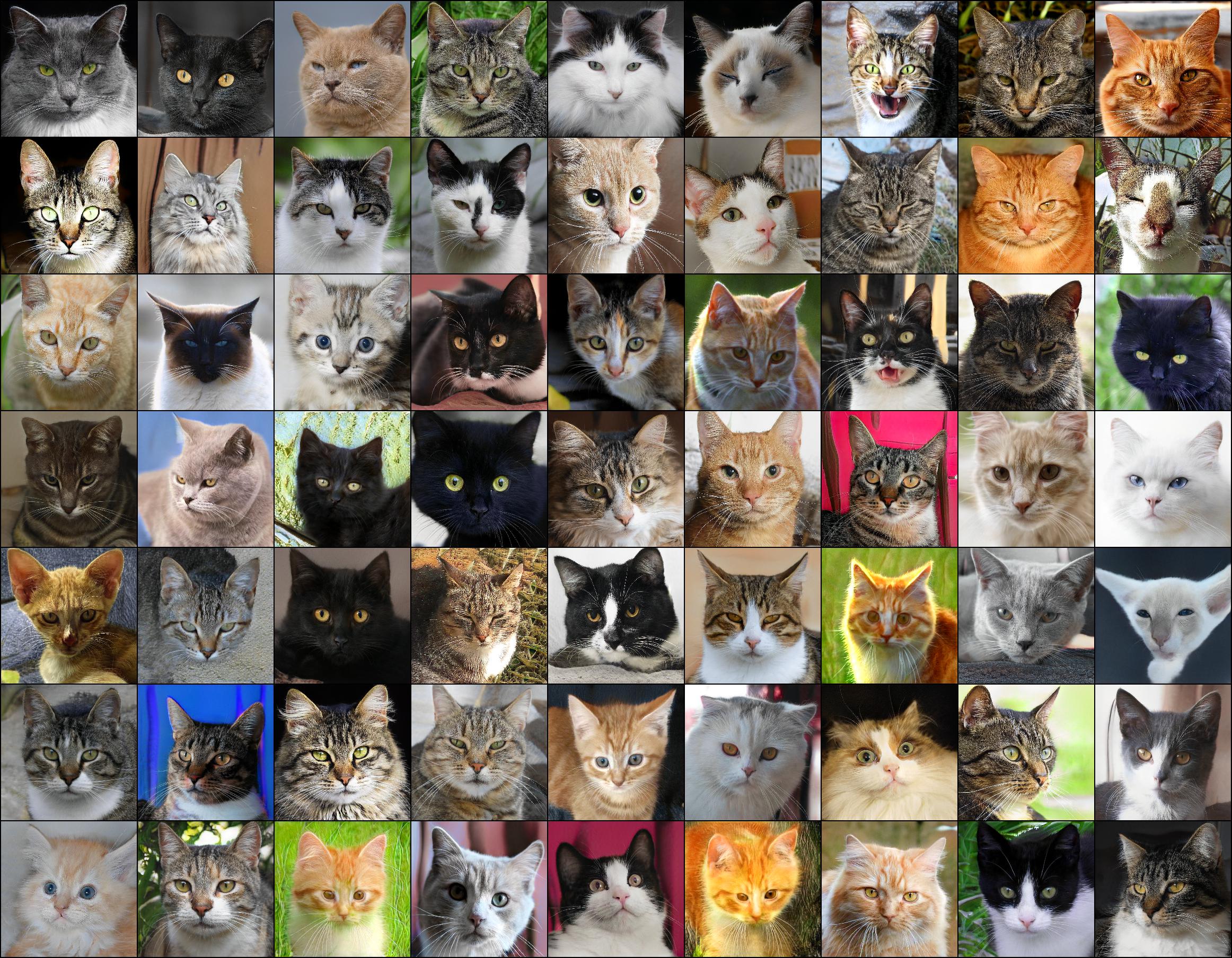}
\end{subfigure}
\par\bigskip
\begin{subfigure}{\linewidth}
\centering
\includegraphics[width=0.75\textwidth]{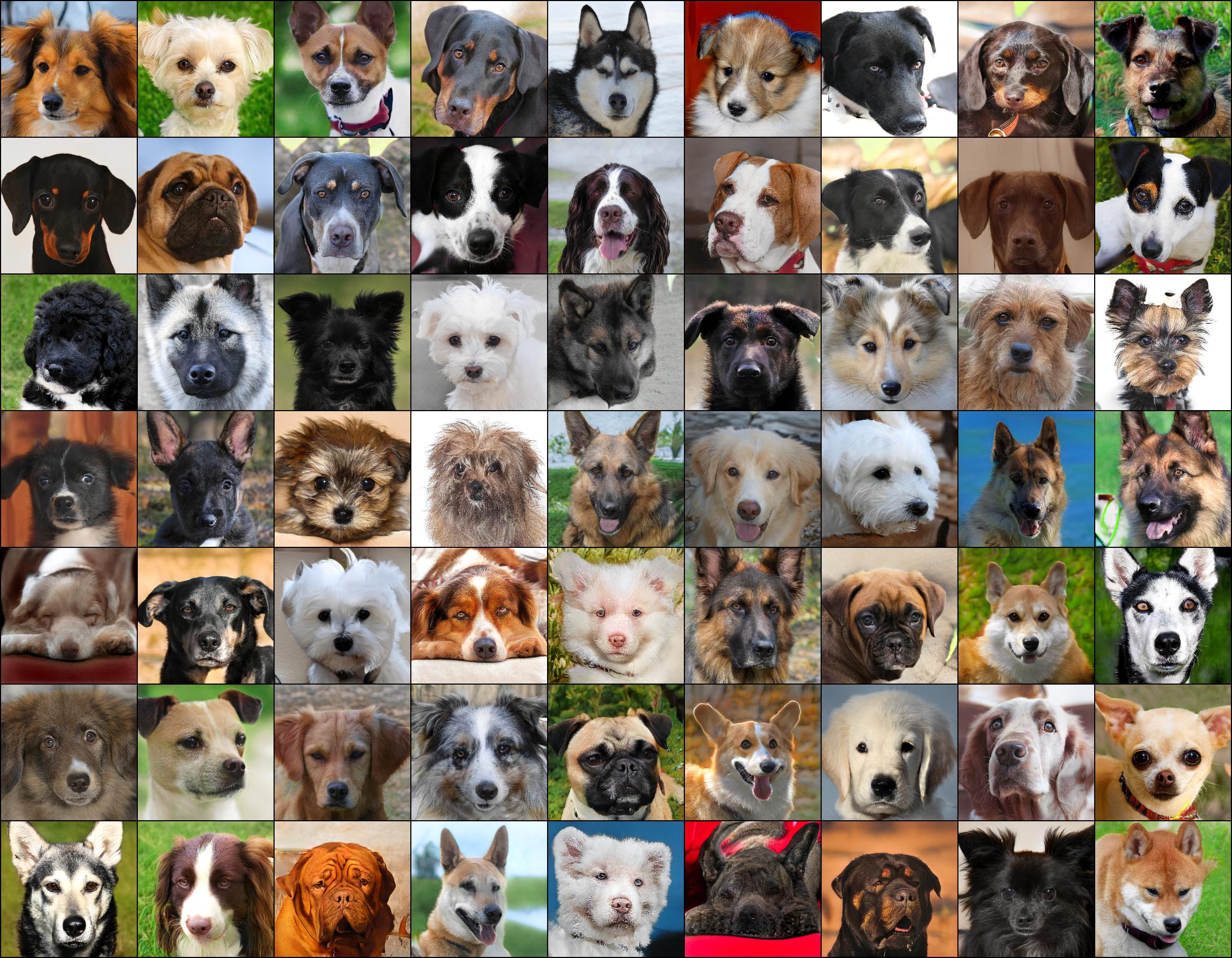}
\end{subfigure}
\caption{Uncurated results on (top) AFHQ Cat and (bottom) AFHQ Dog} 
\label{fig:vis_afhqcat_dog}
\end{figure*}

\begin{figure*}[t]
\centering
\begin{subfigure}{\linewidth}
\centering
\includegraphics[width=0.75\textwidth]{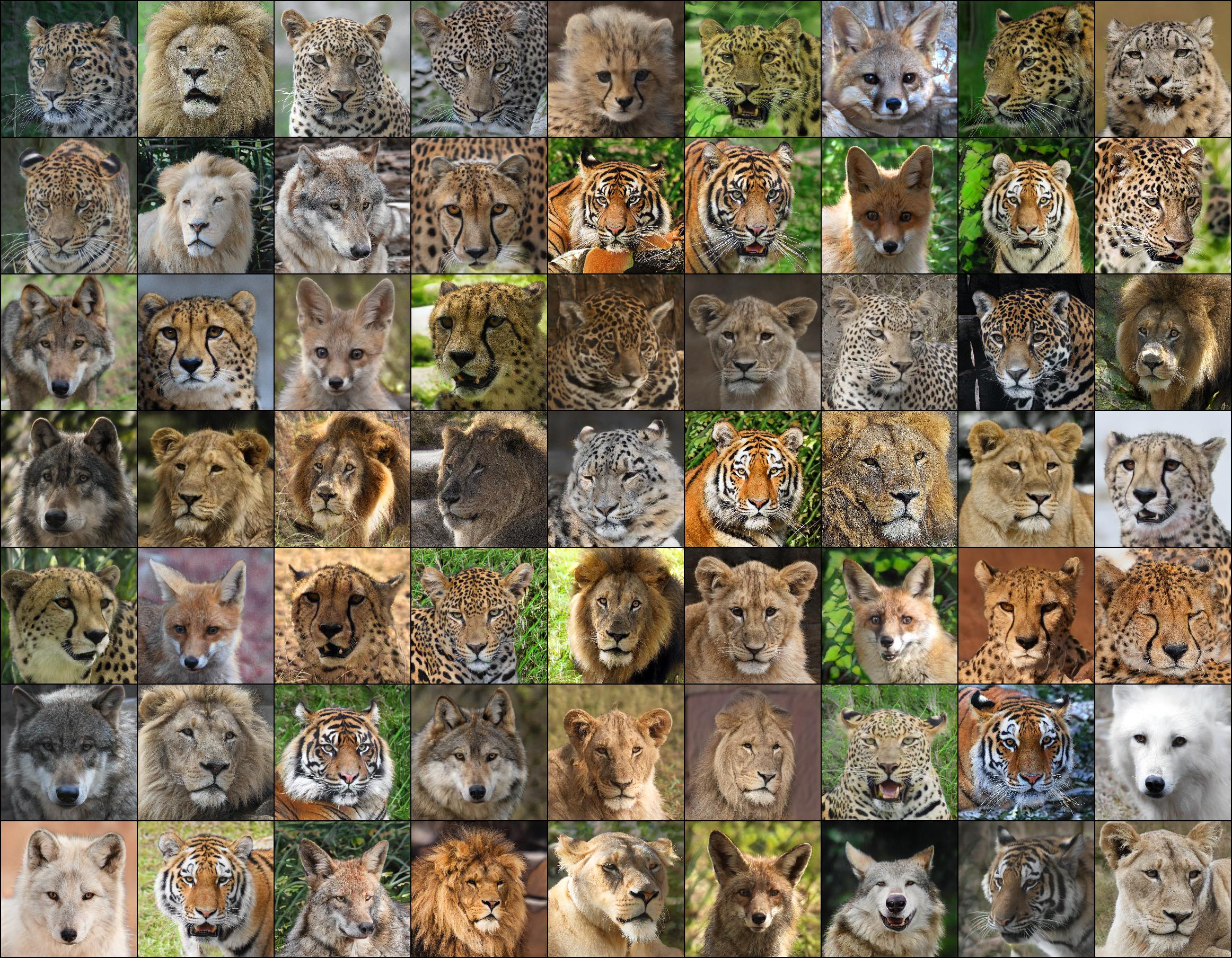}
\end{subfigure}
\par\bigskip
\begin{subfigure}{\linewidth}
\centering
\includegraphics[width=0.75\textwidth]{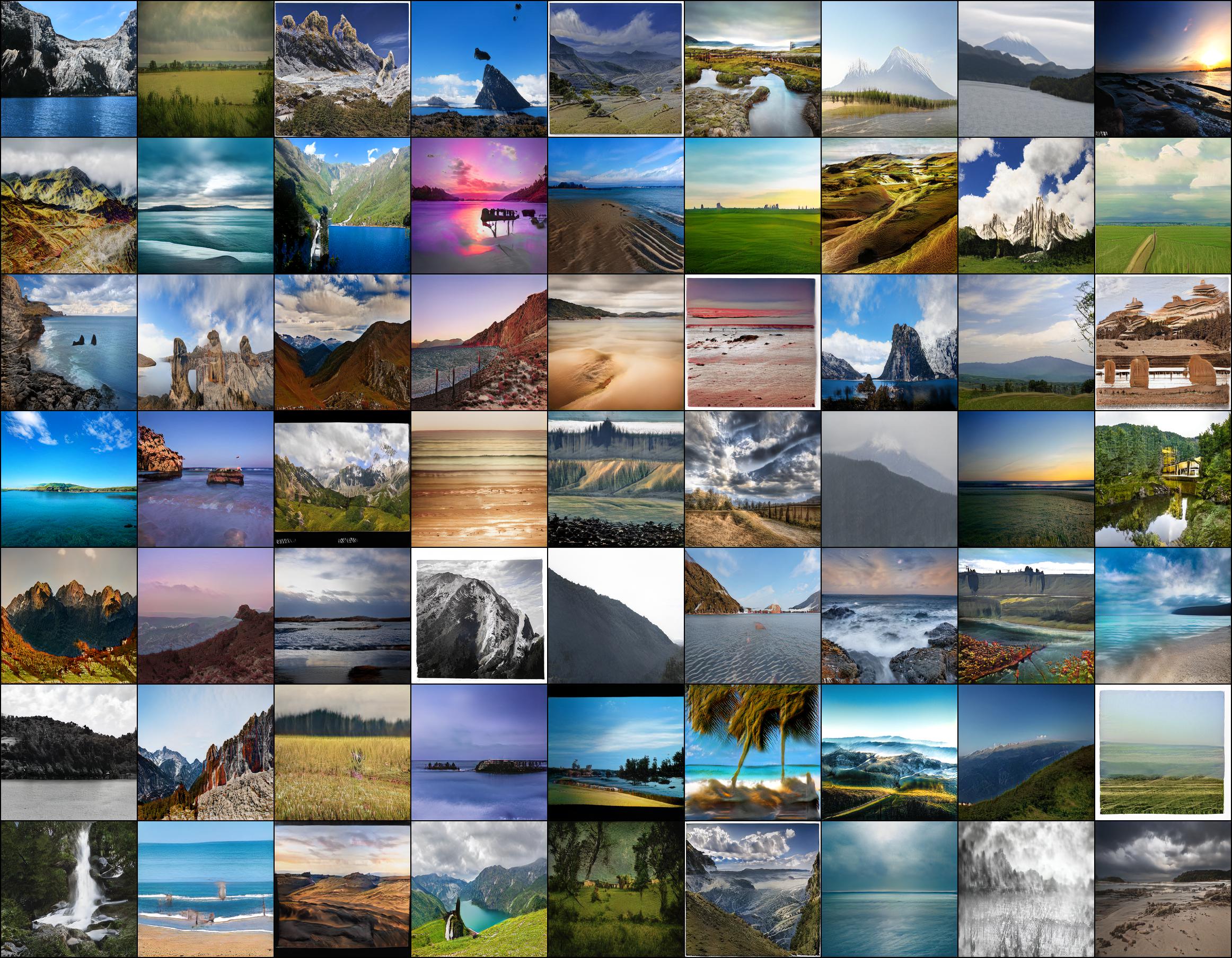}
\end{subfigure}
\caption{Uncurated results on (top) AFHQ Wild and (bottom) Landscape.} 
\label{fig:vis_afhqwild_landscape}
\end{figure*}

\begin{figure*}[t]
\centering
\begin{subfigure}{\linewidth}
\centering
\includegraphics[width=0.75\textwidth]{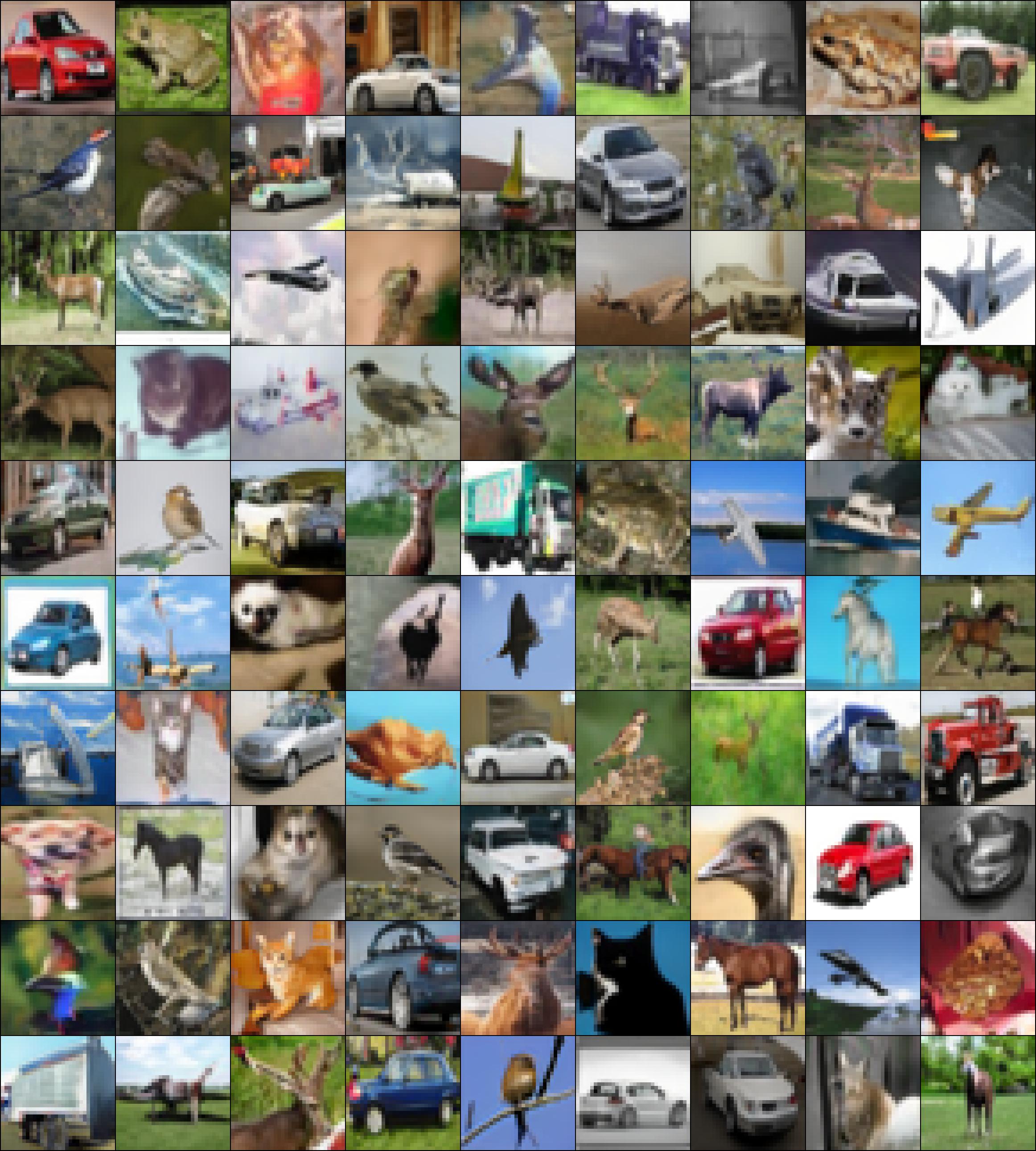}
\end{subfigure}
\caption{Uncurated results on CIFAR-10.} 
\label{fig:vis_cifar10}
\end{figure*}

\begin{figure*}[t]
\centering
\begin{subfigure}{\linewidth}
\centering
\includegraphics[width=0.755\textwidth]{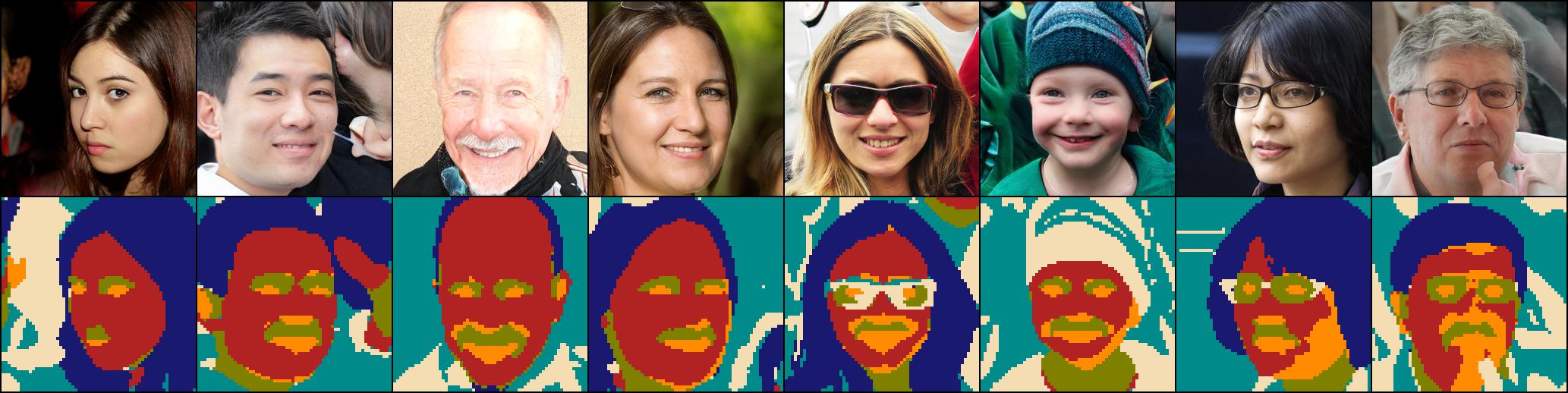}
\includegraphics[width=0.755\textwidth]{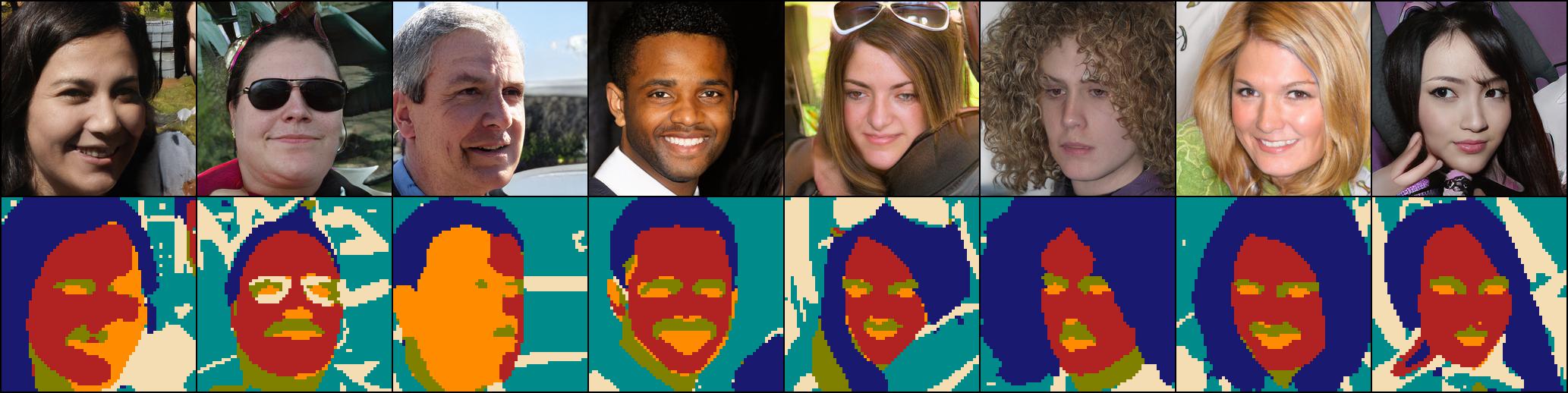}
\caption{Visualization on FFHQ.}
\end{subfigure}
\begin{subfigure}{\linewidth}
\centering
\includegraphics[width=0.755\textwidth]{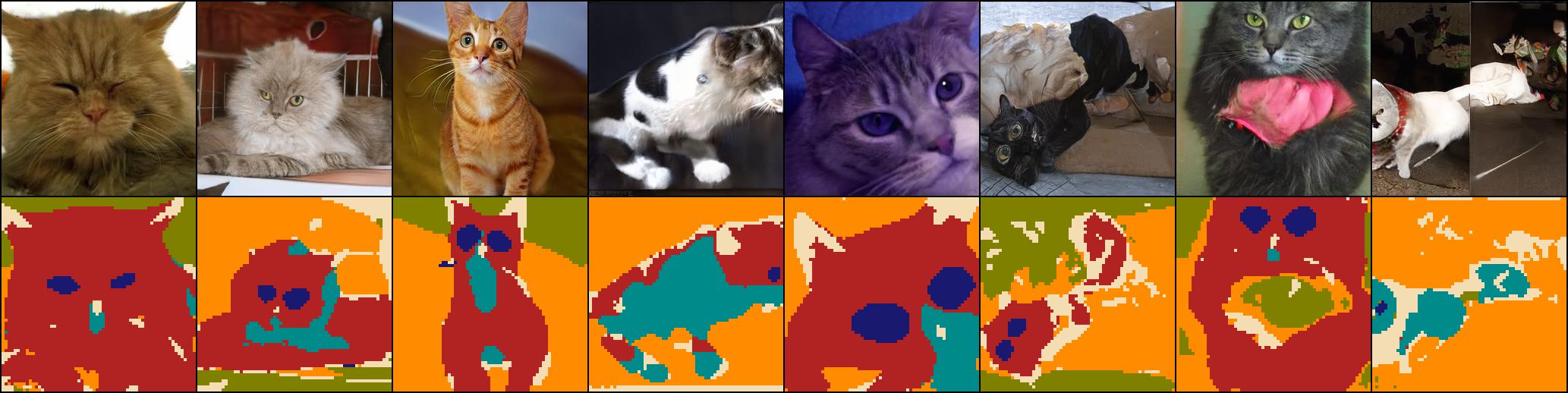}
\includegraphics[width=0.755\textwidth]{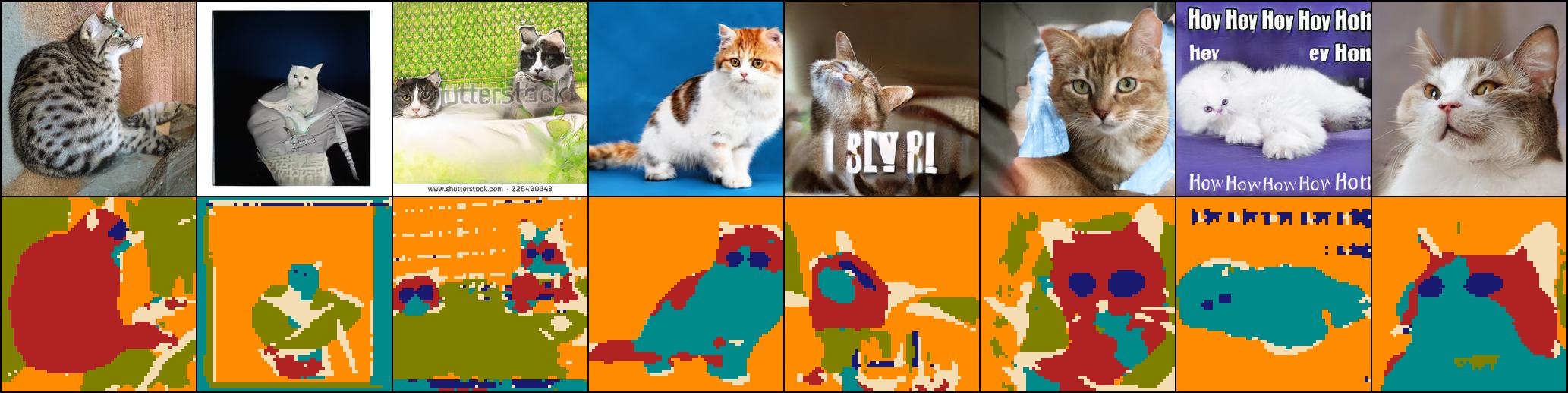}
\caption{Visualization on LSUN Cat.}
\end{subfigure}
\begin{subfigure}{\linewidth}
\centering
\includegraphics[width=0.755\textwidth]{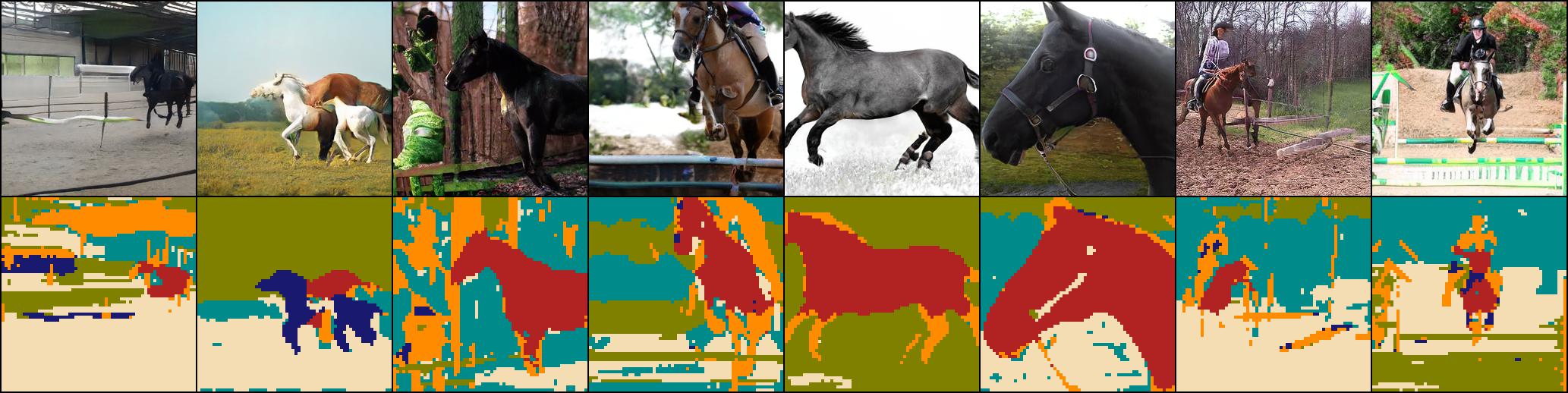}
\includegraphics[width=0.755\textwidth]{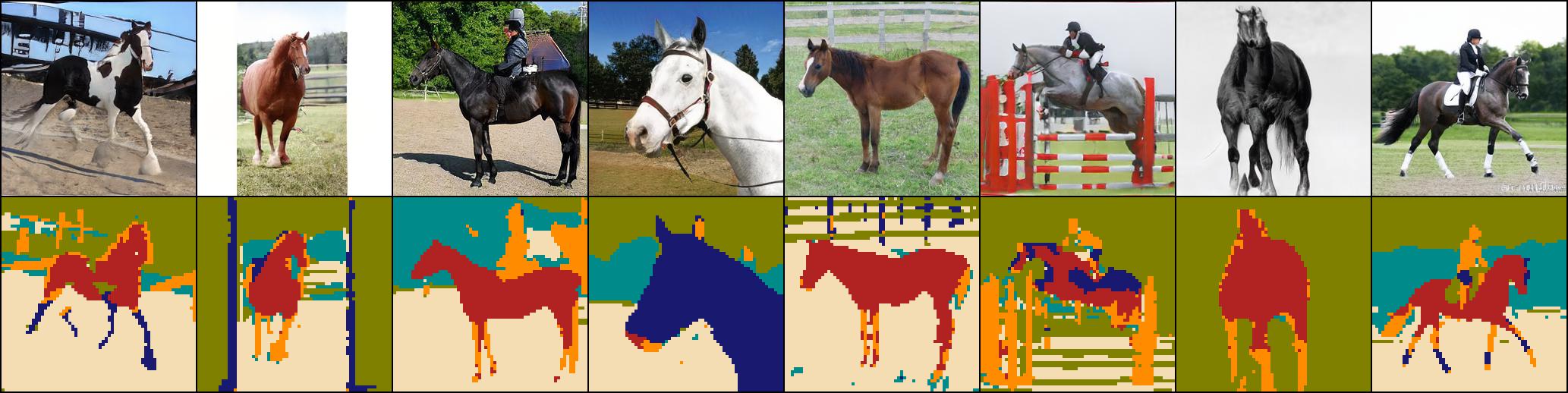}
\caption{Visualization on LSUN Horse.}
\end{subfigure}
\caption{Visualization of fake images and their corresponding generator feature maps on FFHQ, LSUN Cat and LSUN Horse. }
\label{fig:vis_feat1}
\end{figure*}

\begin{figure*}[t]
\centering
\begin{subfigure}{\linewidth}
\centering
\includegraphics[width=0.755\textwidth]{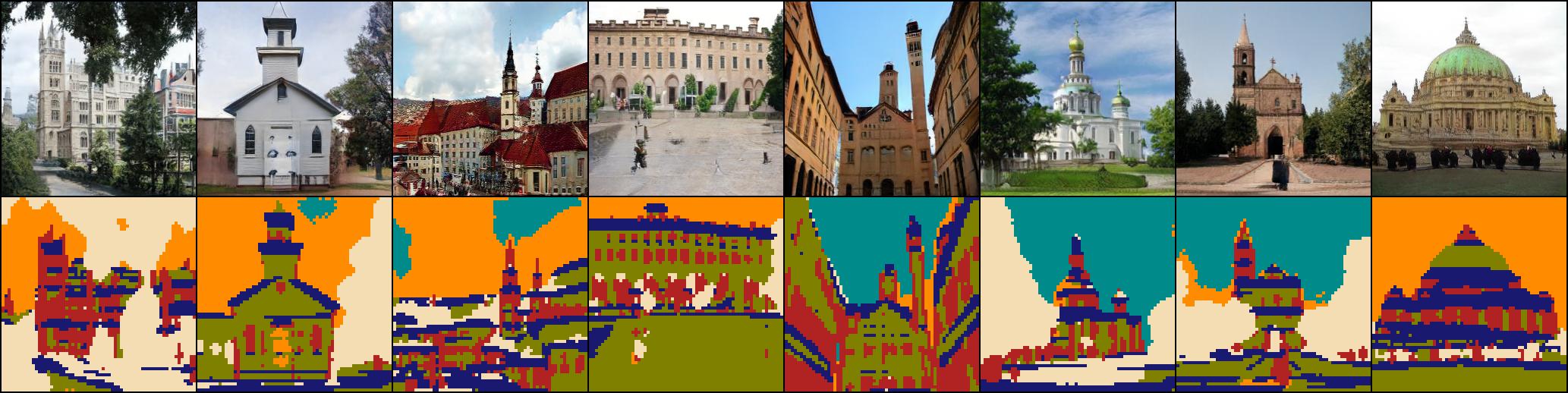}
\includegraphics[width=0.755\textwidth]{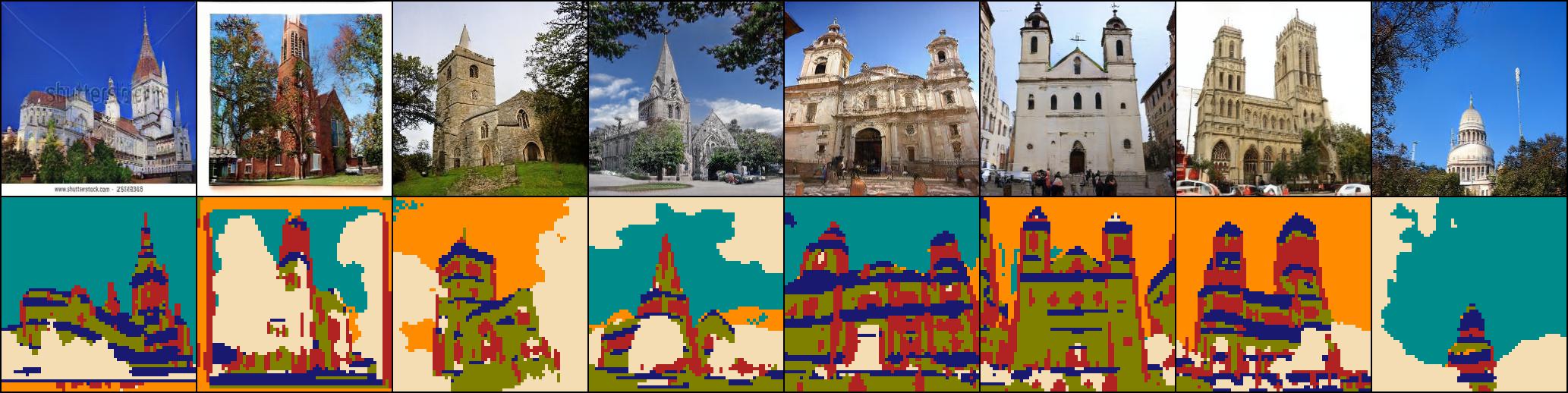}
\caption{Visualization on LSUN Church.}
\end{subfigure}
\begin{subfigure}{\linewidth}
\centering
\includegraphics[width=0.755\textwidth]{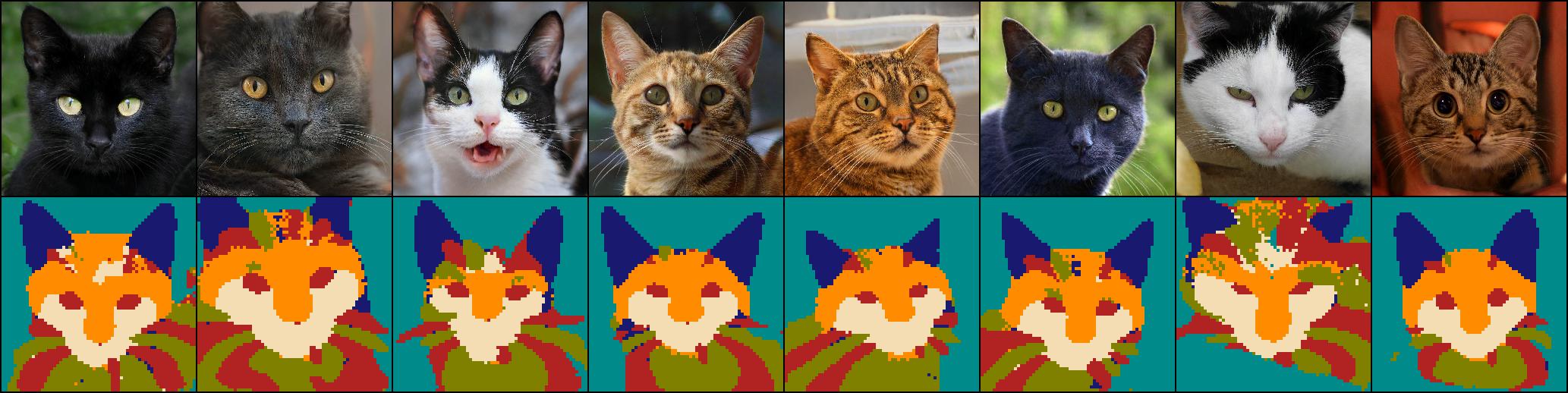}
\includegraphics[width=0.755\textwidth]{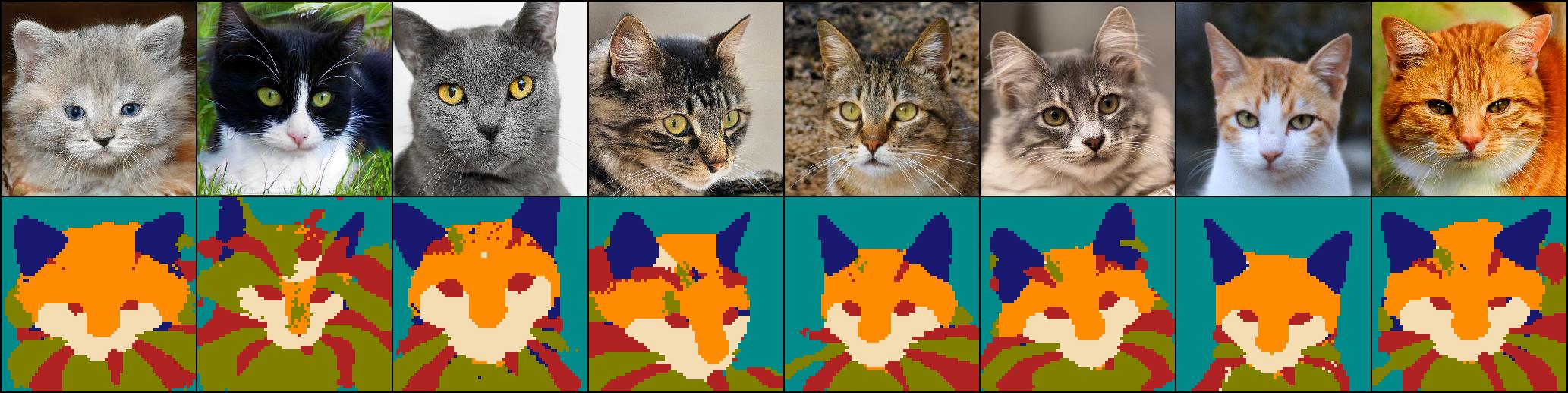}
\caption{Visualization on AFHQ Cat.}
\end{subfigure}
\begin{subfigure}{\linewidth}
\centering
\includegraphics[width=0.755\textwidth]{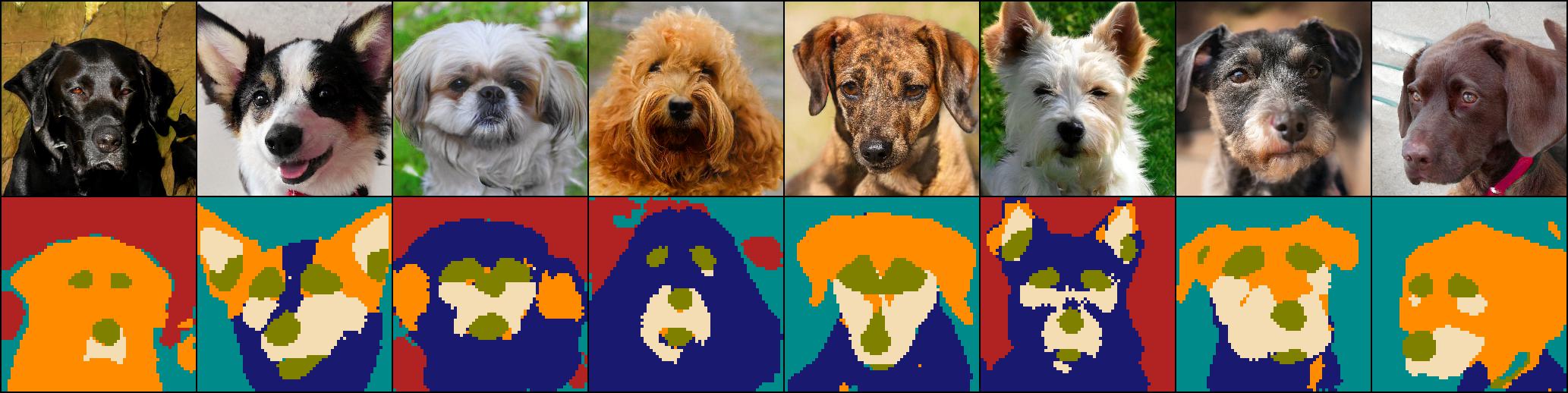}
\includegraphics[width=0.755\textwidth]{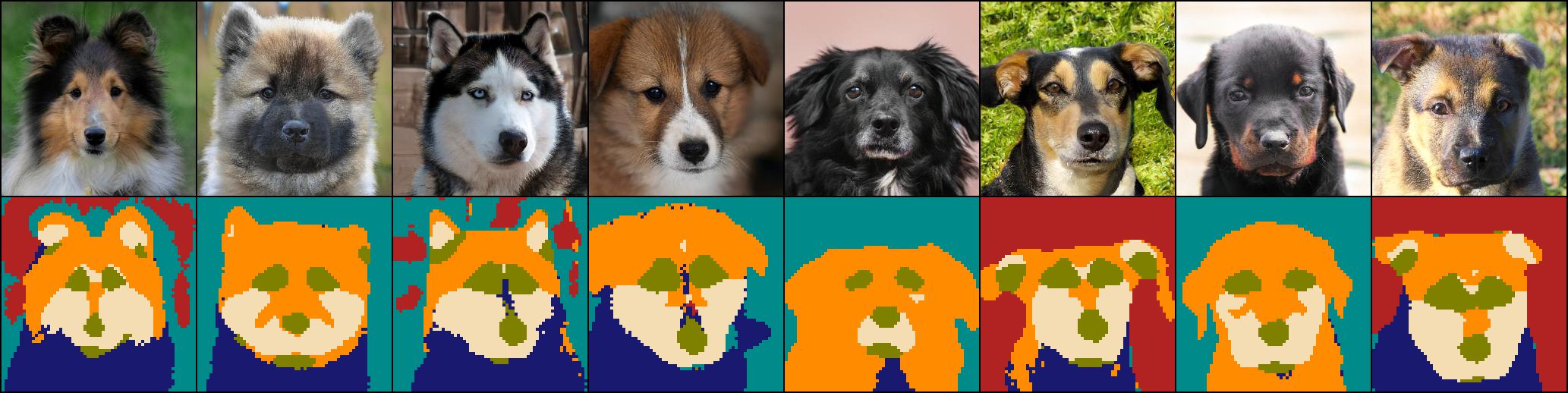}
\caption{Visualization on AFHQ Dog.}
\end{subfigure}
\caption{Visualization of fake images and their corresponding generator feature maps on LSUN Church, AFHQ Cat and AFHQ Dog.}
\label{fig:vis_feat2}
\end{figure*}

\begin{figure*}[t]
\centering
\begin{subfigure}{\linewidth}
\centering
\includegraphics[width=0.755\textwidth]{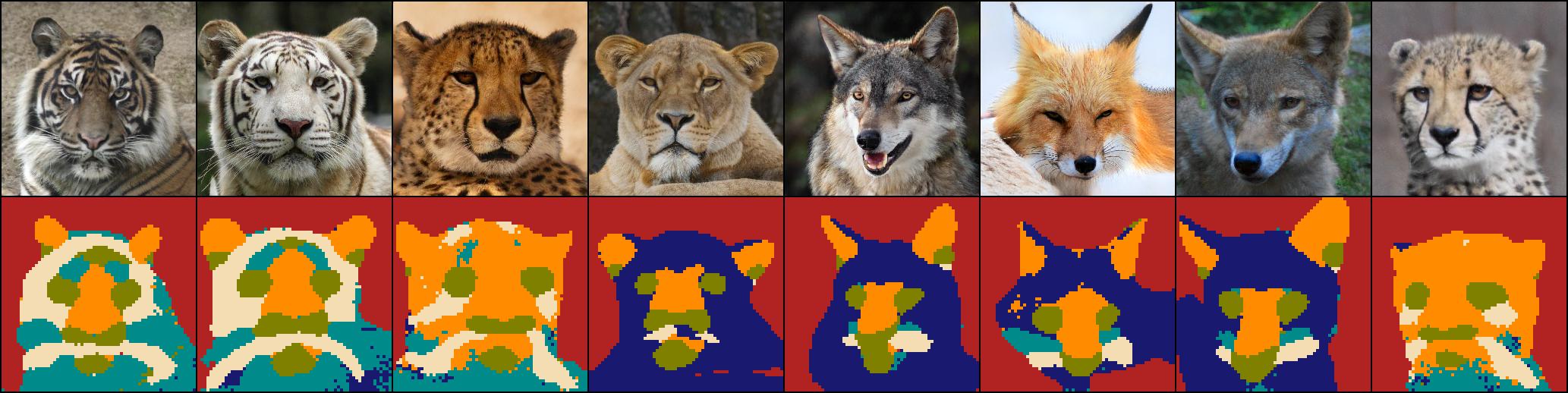}
\includegraphics[width=0.755\textwidth]{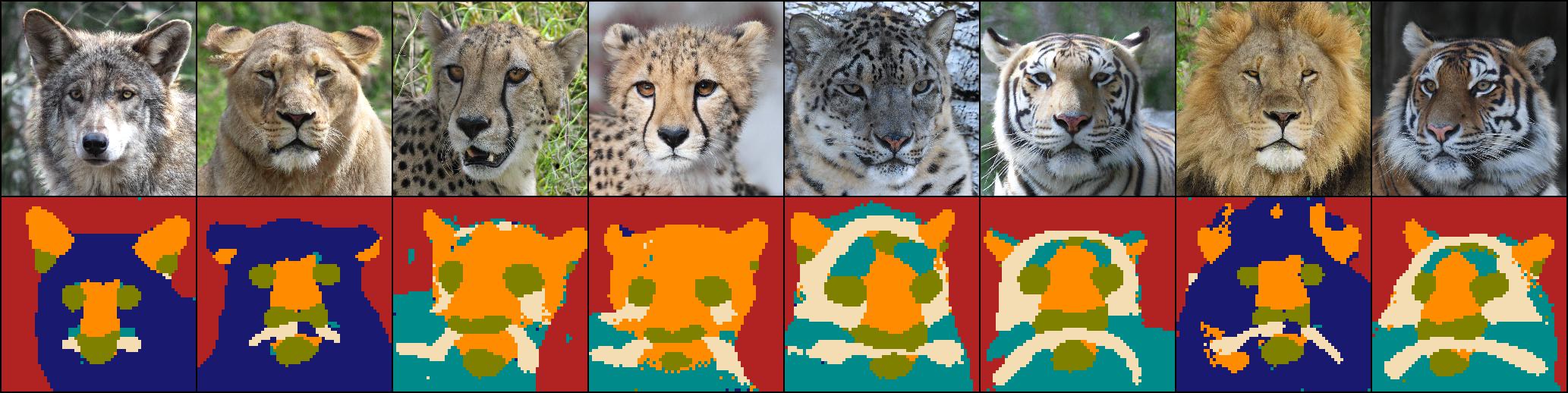}
\caption{Visualization on AFHQ Wild.}
\end{subfigure}
\begin{subfigure}{\linewidth}
\centering
\includegraphics[width=0.755\textwidth]{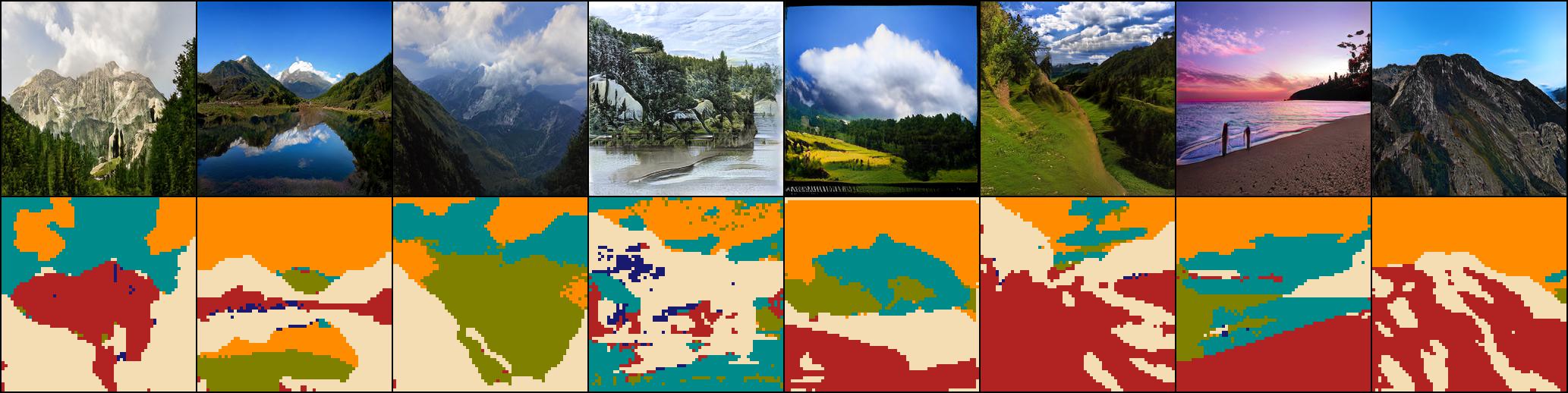}
\includegraphics[width=0.755\textwidth]{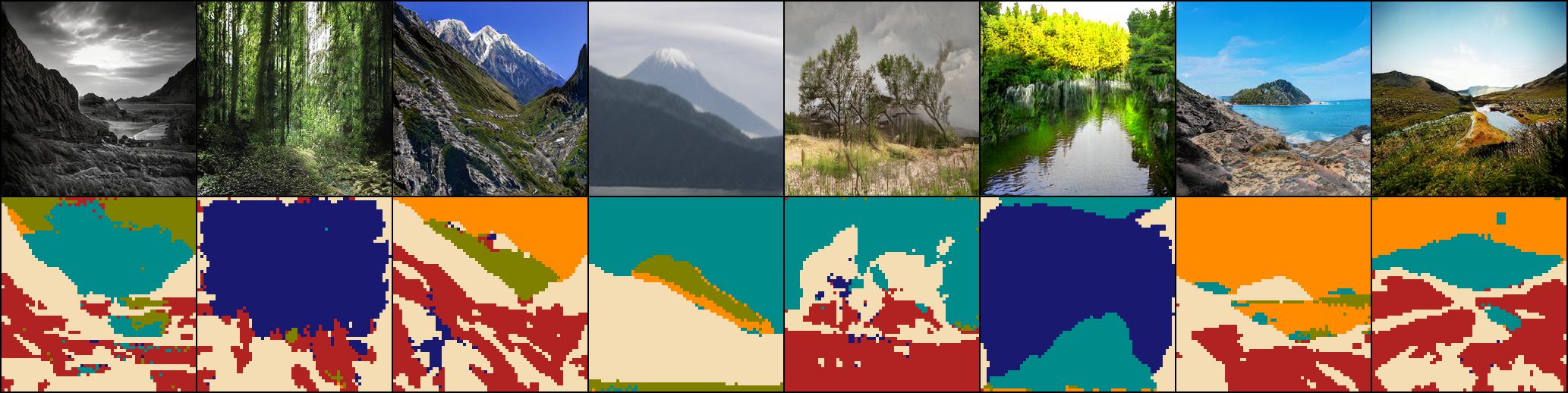}
\caption{Visualization on Landscapes.}
\end{subfigure}
\begin{subfigure}{\linewidth}
\centering
\includegraphics[width=0.755\textwidth]{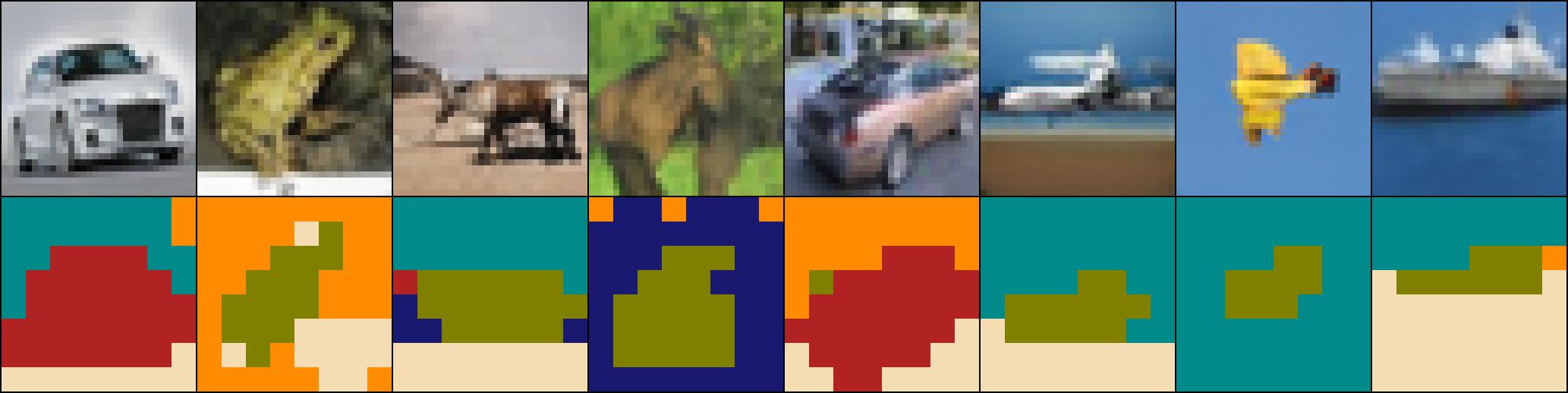}
\includegraphics[width=0.755\textwidth]{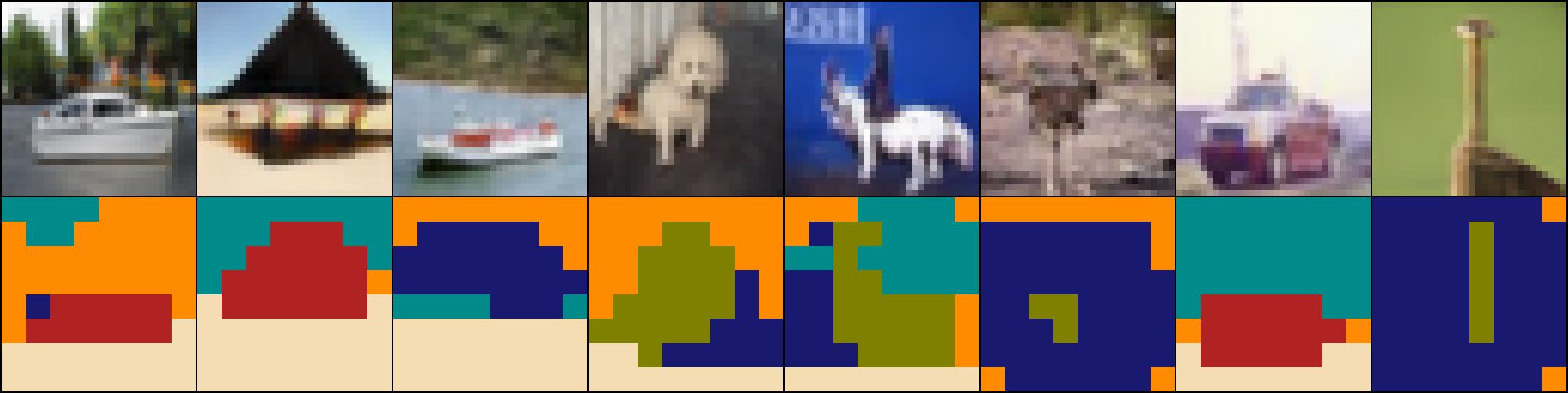}
\caption{Visualization on CIFAR-10.}
\end{subfigure}
\caption{Visualization of fake images and their corresponding generator feature maps on AFHQ Wild, Landscapes and CIFAR-10.}
\label{fig:vis_feat3}
\end{figure*}

We additionally provide results and their corresponding feature visualization for the datasets used in the main paper including CIFAR-10~\cite{krizhevsky2009cifar10}, FFHQ~\cite{karras2019stylegan}, LSUN cat, horse, church~\cite{yu2015lsun}, AFHQ~\cite{choi2020stargan} and Landscapes~\cite{landscape2019kaggle}. From \figref{vis_ffhq_lsuncat} to \figref{vis_cifar10}, we show uncurated random samples generated by our method. For FFHQ and LSUN dataset, we sample from StyleGAN2 trained with GGDR, and we sample from ADA trained with GGDR for the others. From Figure~\ref{fig:vis_feat1} to Figure~\ref{fig:vis_feat3}, we visualize the guidance feature maps from the generator and their corresponding fake images. For feature map visualization, we run $k$-means($k=6$) clustering on 64 samples. All visualized feature map size is $64 \times 64$ except CIFAR-10($8 \times 8$).

\end{document}